\documentclass[10pt,twocolumn,letterpaper]{article}

\usepackage{cvpr}              
\usepackage{comment}
\definecolor{cvprblue}{rgb}{0.21,0.49,0.74}
\definecolor{darkgreen}{rgb}{0.0, 0.75, 0.0}
\definecolor{red}{rgb}{0.9, 0.0, 0.0}
\usepackage{url}
\usepackage{graphicx}
\usepackage{booktabs}
\usepackage{array}     
\usepackage{multirow}  
\usepackage{varioref}
\usepackage{chngcntr}
\usepackage{tabularx}
\newcolumntype{L}{>{\raggedright\arraybackslash}X}
\usepackage{pifont}
\usepackage{tikz}
\usepackage{float}
\usepackage{footnote}
\usepackage[accsupp]{axessibility}

\def\paperID{15414} 
\def\confName{CVPR}
\def\confYear{2025}

\usepackage[pagebackref,breaklinks,colorlinks,allcolors=cvprblue]{hyperref}

\title{EarthDial: Turning Multi-sensory Earth Observations to Interactive Dialogues}

\author{Sagar Soni$^{*1}$, Akshay Dudhane$^{*2}$, Hiyam Debary$^{*1}$, Mustansar Fiaz$^{*1}$, Muhammad Akhtar Munir$^2$ \\ Muhammad Sohail Danish$^2$, Paolo Fraccaro$^1$, Campbell D Watson$^1$, Levente J Klein$^1$ \\ Fahad Shahbaz Khan$^{2,4}$, Salman Khan$^{2,3}$\\
$^1$IBM Research \hspace{2mm} 
$^2$Mohamed bin Zayed University of AI \hspace{2mm} \\$^3$Australian National University \hspace{2mm} $^4$Link\"{o}ping University \hspace{2.5mm}
}

\begin{document}
\maketitle

\begin{abstract}
    Automated analysis of vast Earth observation data via interactive Vision-Language Models (VLMs) can unlock new opportunities for environmental monitoring, disaster response, and {resource management}. Existing generic VLMs do not perform well on Remote Sensing data, while the recent Geo-spatial VLMs remain restricted to a fixed resolution and few sensor modalities.
    In this paper, we introduce EarthDial, a conversational assistant specifically designed for Earth Observation (EO) data, transforming complex, multi-sensory Earth observations into interactive, natural language dialogues.
    EarthDial supports multi-spectral, multi-temporal, and multi-resolution imagery, enabling a wide range of remote sensing tasks, including classification, detection, captioning, question answering, visual reasoning, and visual grounding.
    To achieve this, we introduce an extensive instruction tuning dataset comprising over 11.11M instruction pairs covering RGB, Synthetic Aperture Radar (SAR), and multispectral modalities such as Near-Infrared (NIR) and infrared. Furthermore, EarthDial handles bi-temporal and multi-temporal sequence analysis for applications like change detection.
    Our extensive experimental results on 44 downstream datasets demonstrate that EarthDial outperforms existing generic and domain-specific models, achieving better generalization across various EO tasks. Our source codes and pre-trained models are at \url{https://github.com/hiyamdebary/EarthDial}. 
\end{abstract}

\section{Introduction}
    \footnote{*Equally contributing first authors.}

    Recent advancements in VLMs enable unified visual interpretation, where a single model can perform diverse tasks such as classification, localization, visual question-answering, counting, visual reasoning, and visual grounding \cite{liu2024llavanext, Qwen2VL, chen2024far, achiam2023gpt, team2023gemini, chen2023shikra}.
    However, these generic VLMs do not scale well to Earth Observation (EO) data, which require specialized capabilities to handle the complex geospatial, spectral, and temporal dimensions of remote sensing (RS) data.
    Even state-of-the-art proprietary models like GPT-4V show low accuracies in domain-specific RS data \cite{zhang2024vleobench}, emphasizing the need for EO-specialized VLMs. 

\begin{figure}[t]
    \centering
    \resizebox{\columnwidth}{!}{\input{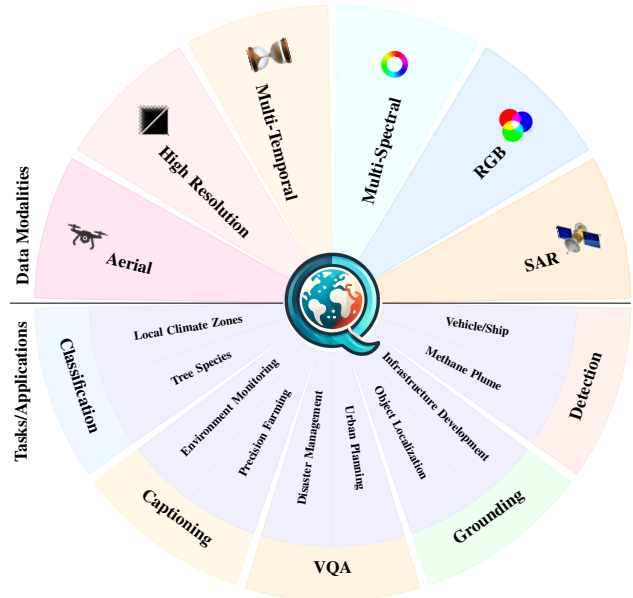}}
    \caption{EarthDial is the first domain-specific VLM for earth observation data that can comprehensively interpret multi-sensor imagery. Specifically, our model covers visible RGB, SAR, multi-temporal, high-res satellite and aerial imagery available in varying spatial resolutions (\emph{top half}). We develop the largest remote sensing image-text instruction dataset with over 11M samples. EarthDial can perform several multimodal understanding tasks: classification, detection, captioning, visual question-answering (VQA), and grounding (\emph{bottom half}).  This unlocks a number of downstream applications where EarthDial shows promising results.}
    \label{fig:data_modalities_tasks}
\end{figure}

\begin{table*}[htpb]
\centering\setlength{\tabcolsep}{2pt}
\resizebox{\textwidth}{!}{
\begin{tabular}{l c c ccc c c c cccccccccccccc}
\hline
& & & & & & \multicolumn{17}{c}{\textbf{Tasks}} \\
\cline{8-23}
\multirow{-2}{*}{\textbf{Dataset}} & \multirow{-2}{*}{\textbf{Type}} & \multirow{-2}{*}{\textbf{\# Samples}} &
\multirow{-2}{*}{\textbf{OS}} & \multirow{-2}{*}{\textbf{MS}} & \multirow{-2}{*}{\textbf{MT}} & \multirow{-2}{*}{\textbf{MR}} & \textbf{IC} & \textbf{RC} & \textbf{VQA} & \textbf{SC}  & \textbf{MLSC}  & \textbf{TSC} & \textbf{OD} & \textbf{VG} & \textbf{DA} & \textbf{BTCD} & \textbf{MTCD} & \textbf{M-TC} & \textbf{UHI} & \textbf{LCZ} &
\textbf{TSC} &\textbf{MPD} \\ \hline

RSICap   \cite{hu2023rsgpt}                                                   & Optical                                                 & 2.6K   & \textcolor{darkgreen}{\checkmark}                                                           & \textcolor{red}{\ding{55}}                                              & \textcolor{red}{\ding{55}} & \textcolor{red}{\ding{55}} & \textcolor{darkgreen}{\checkmark}  & \textcolor{red}{\ding{55}} & \textcolor{red}{\ding{55}} & \textcolor{red}{\ding{55}}                   & \textcolor{red}{\ding{55}}         & \textcolor{red}{\ding{55}}                   & \textcolor{red}{\ding{55}}                                              &  \textcolor{red}{\ding{55}}   & \textcolor{red}{\ding{55}}    & \textcolor{red}{\ding{55}}   & \textcolor{red}{\ding{55}}   & \textcolor{red}{\ding{55}}   & \textcolor{red}{\ding{55}}   & \textcolor{red}{\ding{55}}   & \textcolor{red}{\ding{55}}   & \textcolor{red}{\ding{55}}    \\

VHM \cite{pang2024vhm}                                               & Optical                                                 & 180K  & \textcolor{red}{\ding{55}}                                                            & \textcolor{red}{\ding{55}}                                              & \textcolor{red}{\ding{55}}                                              & \textcolor{red}{\ding{55}}                                              & \textcolor{darkgreen}{\checkmark}  & \textcolor{darkgreen}{\checkmark}   & \textcolor{darkgreen}{\checkmark}   & \textcolor{darkgreen}{\checkmark}  & \textcolor{darkgreen}{\checkmark}         & \textcolor{red}{\ding{55}}  & \textcolor{darkgreen}{\checkmark}   & \textcolor{darkgreen}{\checkmark}  & \textcolor{red}{\ding{55}}   & \textcolor{red}{\ding{55}}   & \textcolor{red}{\ding{55}}   & \textcolor{red}{\ding{55}} & \textcolor{red}{\ding{55}} & \textcolor{red}{\ding{55}} & \textcolor{red}{\ding{55}}   & \textcolor{red}{\ding{55}}   \\

VRSBench   \cite{li2024vrsbench}                                                   & Optical                                                 & 205K   & \textcolor{darkgreen}{\checkmark}                                                           & \textcolor{red}{\ding{55}}                                              & \textcolor{red}{\ding{55}} & \textcolor{red}{\ding{55}} & \textcolor{darkgreen}{\checkmark}  & \textcolor{darkgreen}{\checkmark} & \textcolor{darkgreen}{\checkmark} & \textcolor{red}{\ding{55}}                   & \textcolor{red}{\ding{55}}         & \textcolor{red}{\ding{55}}                   & \textcolor{darkgreen}{\checkmark}                                             &  \textcolor{red}{\ding{55}}   & \textcolor{red}{\ding{55}}    & \textcolor{red}{\ding{55}}   & \textcolor{red}{\ding{55}}   & \textcolor{darkgreen}{\checkmark}   & \textcolor{red}{\ding{55}}   & \textcolor{red}{\ding{55}}   & \textcolor{red}{\ding{55}}   & \textcolor{red}{\ding{55}}    \\

GeoChat  \cite{kuckreja2024geochat}                                                   & Optical                                                 & 380K     & \textcolor{darkgreen}{\checkmark}                                                        & \textcolor{red}{\ding{55}}                                              & \textcolor{red}{\ding{55}}                                        & \textcolor{red}{\ding{55}}                                              & \textcolor{darkgreen}{\checkmark}  & \textcolor{darkgreen}{\checkmark}  & \textcolor{darkgreen}{\checkmark}   & \textcolor{darkgreen}{\checkmark}  & \textcolor{red}{\ding{55}}         & \textcolor{red}{\ding{55}}  & \textcolor{darkgreen}{\checkmark}  & \textcolor{darkgreen}{\checkmark}   & \textcolor{red}{\ding{55}} & \textcolor{red}{\ding{55}} & \textcolor{red}{\ding{55}} & \textcolor{darkgreen}{\checkmark}       & \textcolor{red}{\ding{55}}   & \textcolor{red}{\ding{55}}   & \textcolor{red}{\ding{55}}   & \textcolor{red}{\ding{55}}    \\
MMRS   \cite{zhang2024earthgpt}                                                       & Optical, SAR, IR                                         & 1.01M   & \textcolor{darkgreen}{\checkmark}                                                   & \textcolor{red}{\ding{55}}                                              & \textcolor{red}{\ding{55}}                                               & \textcolor{red}{\ding{55}}                                              & \textcolor{darkgreen}{\checkmark}  & \textcolor{darkgreen}{\checkmark}  & \textcolor{darkgreen}{\checkmark}   & \textcolor{darkgreen}{\checkmark}   & \textcolor{red}{\ding{55}}         & \textcolor{red}{\ding{55}}  & \textcolor{darkgreen}{\checkmark}  & \textcolor{darkgreen}{\checkmark}  & \textcolor{red}{\ding{55}}   & \textcolor{red}{\ding{55}}  & \textcolor{red}{\ding{55}} & \textcolor{darkgreen}{\checkmark}  & \textcolor{red}{\ding{55}} & \textcolor{red}{\ding{55}} & \textcolor{red}{\ding{55}}   & \textcolor{red}{\ding{55}}    \\
SkyEye-968k    \cite{zhan2024skyeyegpt}                                             & Optical                                                 &  0.97M         & \textcolor{darkgreen}{\checkmark}                                                 & \textcolor{red}{\ding{55}}                                              &  \textcolor{red}{\ding{55}}                                              & \textcolor{red}{\ding{55}}                                              & \textcolor{darkgreen}{\checkmark}  & \textcolor{red}{\ding{55}}   & \textcolor{darkgreen}{\checkmark}   & \textcolor{darkgreen}{\checkmark}  & \textcolor{red}{\ding{55}}         & \textcolor{red}{\ding{55}}  & \textcolor{darkgreen}{\checkmark} & \textcolor{darkgreen}{\checkmark} & \textcolor{red}{\ding{55}}  & \textcolor{red}{\ding{55}}   & \textcolor{red}{\ding{55}}   & \textcolor{darkgreen}{\checkmark}  & \textcolor{red}{\ding{55}}  & \textcolor{red}{\ding{55}}  & \textcolor{red}{\ding{55}}   & \textcolor{red}{\ding{55}}   \\
LHRS-Instruct    \cite{muhtar2024lhrs}                                           & Optical                                                 & 81K          & \textcolor{darkgreen}{\checkmark}                                                 & \textcolor{red}{\ding{55}}                                           & \textcolor{red}{\ding{55}}                                              & \textcolor{red}{\ding{55}}                                              & \textcolor{darkgreen}{\checkmark}  & \textcolor{red}{\ding{55}}   & \textcolor{darkgreen}{\checkmark}   & \textcolor{darkgreen}{\checkmark}   & \textcolor{red}{\ding{55}}         & \textcolor{red}{\ding{55}} & \textcolor{red}{\ding{55}}   & \textcolor{darkgreen}{\checkmark}  & \textcolor{red}{\ding{55}}   & \textcolor{red}{\ding{55}}   &  \textcolor{red}{\ding{55}} & \textcolor{darkgreen}{\checkmark}   & \textcolor{red}{\ding{55}}  & \textcolor{red}{\ding{55}} & \textcolor{red}{\ding{55}}   & \textcolor{red}{\ding{55}}   \\
FIT-RS    \cite{luo2024skysensegpt}                                                 & Optical                                                 & 1.8M     & \textcolor{darkgreen}{\checkmark}                                                   & \textcolor{red}{\ding{55}}                                             & \textcolor{red}{\ding{55}}                                            & \textcolor{red}{\ding{55}}                                              & \textcolor{darkgreen}{\checkmark}  & \textcolor{darkgreen}{\checkmark}  & \textcolor{darkgreen}{\checkmark}   & \textcolor{darkgreen}{\checkmark}   & \textcolor{red}{\ding{55}}         & \textcolor{red}{\ding{55}}  & \textcolor{darkgreen}{\checkmark}  & \textcolor{darkgreen}{\checkmark}  & \textcolor{red}{\ding{55}}  &\textcolor{red}{\ding{55}}  & \textcolor{red}{\ding{55}}  & \textcolor{darkgreen}{\checkmark}  & \textcolor{red}{\ding{55}} & \textcolor{red}{\ding{55}}   & \textcolor{red}{\ding{55}}   &  \textcolor{red}{\ding{55}}    \\
\midrule
\textbf{EarthDial-Instruct (ours)}                                         & Optical, SAR, S2, IR, NAIP                               & 11.11M                                               & \textcolor{darkgreen}{\textcolor{darkgreen}{\checkmark}}         & \textcolor{darkgreen}{\checkmark}                                            & \textcolor{darkgreen}{\checkmark}                                            & \textcolor{darkgreen}{\checkmark}                                            & \textcolor{darkgreen}{\checkmark}  & \textcolor{darkgreen}{\checkmark}  & \textcolor{darkgreen}{\checkmark}   & \textcolor{darkgreen}{\checkmark}  & \textcolor{darkgreen}{\checkmark}  & \textcolor{darkgreen}{\checkmark}  & \textcolor{darkgreen}{\checkmark}  & \textcolor{darkgreen}{\checkmark}  & \textcolor{darkgreen}{\checkmark} & \textcolor{darkgreen}{\checkmark}  & \textcolor{darkgreen}{\checkmark}  & \textcolor{darkgreen}{\checkmark}  & \textcolor{darkgreen}{\checkmark}  & \textcolor{darkgreen}{\checkmark}  & \textcolor{darkgreen}{\checkmark}  & \textcolor{darkgreen}{\checkmark}   \\
\hline
\end{tabular}
}\vspace{-0.5em}
\caption{Overview of RS VLM datasets, their types, and supported tasks. OS: Open-Source, MS: Multi-Spectral, MT: Multi-Temporal, MR: Multi-Resolution. Tasks: IC: Image Captioning, RC: Region Captioning, VQA: Visual Question Answering, SC: Scene Classification, MLSC: Multi-Label Scene Classification,  TSC: Temporal Scene Classification, OD: Object Detection, VG: Visual Grounding, DA: Disaster Assessment, BTCD: Bi-Temporal Change Detection, MTCD: Multi-Temporal Change Detection, M-TC: Multi-Task Conversation, MPD: Methane Plume Detection, UHI: Urban Heat Island, TSC: Tree Species Classification, LCZ: Local Climate Zones. Our dataset is 6$\times$ larger and covers diverse sensing modalities and provides annotations for a rich set of downstream tasks. }
\vspace{-0.5em}
\label{tab:dataset-tasks}
\end{table*}

    Recently, domain-specific VLMs have been developed to understand EO data using generative multimodal models.
    RS-GPT fine-tuned the MiniGPT-4 model on 2.5k remote sensing instructions \cite{hu2023rsgpt}. 
    GeoChat represents an initial effort capable of performing image and region-level understanding as well as visual grounding in high-resolution remote sensing images \cite{kuckreja2024geochat}.
    Several approaches have focused on data scaling; e.g., LHRS-Bot uses crowd-sourced labels from OpenStreetMap to obtain 1.15M RS image-text pairs for multimodal alignment \cite{muhtar2024lhrs}. SkyEyeGPT curates a 968K sample instruction-following dataset for remote sensing conversational tasks \cite{zhan2024skyeyegpt}.
    However, these efforts are limited in high-resolution image processing and do not support multi-spectral, multi-temporal analysis.

    In this work, we present EarthDial, aiming to develop the first unified model that can cohesively process multi-resolution, multi-spectral, and multi-temporal remote sensing imagery to unlock numerous downstream tasks. 
    In all the above modalities, EarthDial can perform diverse tasks, including classification, object/change detection, question-answering, image and region captioning, and visual grounding.
    To achieve this goal, we propose the most extensive instruction tuning dataset to date, with over 11.11M instructions covering visible imagery with varying resolutions, SAR, and multispectral modalities, including NIR and infrared.
    Furthermore, EarthDial can handle bi-temporal and multi-temporal sequence analysis for applications such as change detection and temporal sequence classification.

\noindent Our main contributions are as follows:
\begin{itemize}
    \item We propose EarthDial, a conversational VLM capable of processing multi-spectral, multi-temporal, and multi-resolution remote sensing imagery with natural text queries, addressing a wide range of EO tasks.
    \item We introduce the largest instruction tuning dataset for remote sensing, comprising over 11.11M instruction pairs across various modalities, enhancing the model's understanding and generalization capabilities.
    \item  Experimental results demonstrate that EarthDial performs well in comparison to existing domain-specific VLMs, achieving higher accuracies and better generalization across 44 downstream EO tasks.
\end{itemize}

\begin{figure*}[t]
\centering
 \includegraphics[width=0.95\linewidth]{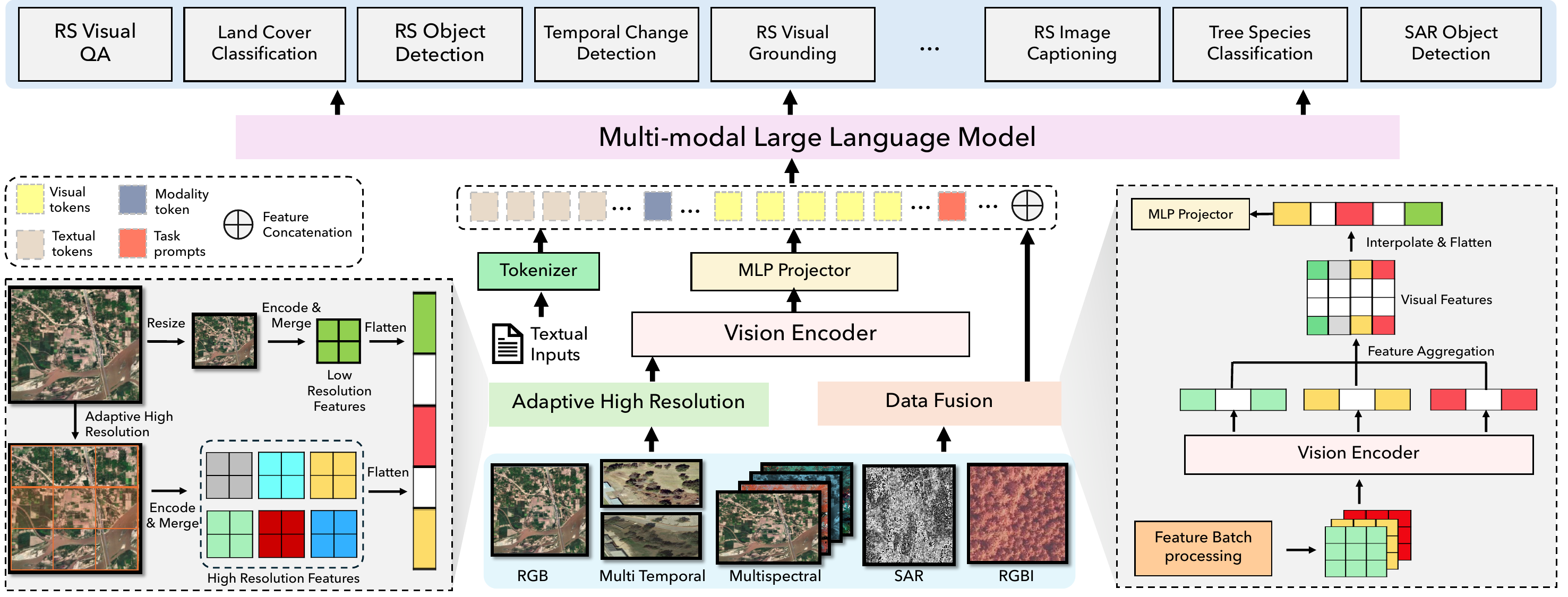} 
\caption{EarthDial Architecture: The model can take a diverse set of inputs ranging from RGB to multi-spectral and time-series images. Multi-resolution inputs are converted to tokens based on an adaptive high-resolution block \cite{chen2024far} that includes both local and global features. The multi-channel inputs (multi-spectral/temporal) are converted to tokens via the data fusion block, which aggregates features across all channels. The resulting visual tokens are mapped to LLM input space using MLP projectors and concatenated with the textual inputs. We use special task and modality tokens to distinguish between several input modalities and downstream tasks (Table~\ref{tab:vqa_samples}). The LLM is trained with multimodal inputs to perform a number of downstream tasks, ranging from VQA to detection, grounding and change detection. }
 \label{fig:overall_arch}
\end{figure*}

\section{Related Work}
\label{sec:related_work}

\noindent\textbf{Generic Vision-Language Models (VLMs):} 
The development of generic VLMs like VisualGPT \cite{chen2022visualgpt}, BLIP \cite{li2023blip}, Flamingo \cite{alayrac2022flamingo}, and Kosmos \cite{huang2023language} has enabled advancements in showcasing multi-modal understanding by aligning visual and language data for diverse applications. Devoted efforts from researchers enabled the VLMs to perform a range of tasks, for example, OCR to diagram and infographics understanding to video analysis within a unified model \cite{Qwen2VL, chen2024far, li2025llama, li2024llava}. The continuous progress enabled the alignment of additional modalities such as audio, video, 3D point clouds \cite{deng2024vg4d, wang2024comprehensive}, audio-video grounding tasks \cite{chowdhury2024meerkat}, 3D visual grounding \cite{xu2024vlm} as well as in fields such as LIDAR \cite{zhou2024openannotate3d, liao2024vlm2scene} and robotics \cite{gao2024physically, wang2024solving}. 
Nevertheless, they struggle with the unique contextual complexities of remote sensing (RS) data, which requires specialized alignment for geospatial, spectral, and temporal information.

\noindent\textbf{Geospatial VLMs:}
Recently, various efforts have been devoted towards domain-specific RS vision-language understanding to address the limitations of general VLMs \cite{liu2024remoteclip, hu2023rsgpt, kuckreja2024geochat, muhtar2024lhrs, zhan2024skyeyegpt, zhang2024earthgpt}. 
RemoteCLIP \cite{liu2024remoteclip} employs contrastive learning over  RS image-text pairs, illustrating the zero-shot classification and image-text retrieval capabilities.
RS-GPT \cite{hu2023rsgpt} fine-tuned over EVA-CLIP and Vicuna LLM demonstrates image captioning and VQA abilities while struggling over detection and visual grounding tasks. 
GeoChat \cite{kuckreja2024geochat}, LHRS-Bot \cite{muhtar2024lhrs} and SkyEyeGPT \cite{zhan2024skyeyegpt} extend their capabilities to resolve multiple tasks such as region-level understanding as well as visual grounding in high-resolution RS images. However, these models do not cover multi-spectral and temporal modalities. 
More recently, EarthGPT \cite{zhang2024earthgpt} introduces the MMRS1M dataset to integrate optical, SAR, and infrared modalities, advancing multi-sensor RS comprehension. 
However, they do not cover other multi-spectral inputs and lack generalization to multi-temporal and varying resolution inputs.

\noindent\textbf{RS Instruction Datasets:}
A large number of instruction-following datasets have been introduced to train the RS-VLMs effectively. For example,  GeoChat-Instruct \cite{kuckreja2024geochat}, SkyEye-968k \cite{zhan2024skyeyegpt}, and RS5M \cite{zhang2024rs5m} provide extensive RS image-text pairs, supporting instruction-based VLM training on optical data. These existing datasets limit the model's capabilities across different sensor modalities.
On the other hand,  MLLMs \cite{zhang2024earthgpt}, provide image-text pairs from optical, SAR, and infrared images. 
Regardless of the large scale, the aforementioned datasets often lack diverse RS applications across various modalities, including multi-resolution, multi-spectral, and multi-temporal RS sensor data.
The recently introduced dataset aims to enhance content richness with factual and deceptive questions, enabling VLMs to address a wider range of tasks in the RS domain \cite{pang2024vhm}.
However, there is no large-scale unified instruction-following dataset that can encapsulate the distinctive contextual
complexities of diverse RS applications across different modalities, multi-resolution, multi-spectral, and multi-temporal RS sensor data. Our work is an effort to bridge this gap with an 11M instruction set to seamlessly integrate multiple earth observation modalities covering diverse spectral and with time-series imaging data for diverse RS applications.

\section{EarthDial}
\label{sec:geofusion}

Our goal is to develop a domain-specialized  VLM that can handle complex geospatial, spectral, and temporal dimensions unique to RS imagery. 
As described above, the existing general and geospatial VLMs lack in understanding high-resolution, multi-spectral, and multi-temporal RS imagery. 
To bridge this gap, we propose a comprehensive large-scale instruction tuning dataset for RS domain with over 11M instruction, covering diverse resolutions and geographical locations.
Building on this dataset, we propose EarthDial, the first unified model capable of processing multi-resolution, multi-spectral, and multi-temporal RS data across a variety of tasks, from classification and visual grounding to change detection.

EarthDial leverages state-of-the-art vision-language models (VLMs) for natural images and provides a multi-stage finetuning recipe to progressively expand model capabilities. 
Our model architecture builds on InternVL \cite{chen2024far,chen2024internvl} with specific modifications to enable multi-spectral and multi-temporal processing (Sec.~\ref{sec:model_arch}).  
We proposed a three-stage model training process to enhance the model's capabilities across multiresolution, multispectral, and multitemporal datasets.
We first pretrain with remote sensing datasets, focusing on adapting state-of-the-art VLMs for EO-specific dialogues.
In the next stage, output of pretrained encoders and LLM are adapted using RGB and temporal imagery for downstream tasks.
Finally, an extended finetuning stage is specifically designed to improve its performance with multispectral and synthetic aperture radar (SAR) datasets to broadly cover additional applications. 
Next, we explain our model design in detail.

    \subsection{Model architecture}\label{sec:model_arch}

\begin{figure}[t]
\centering
 \includegraphics[width=0.9\linewidth]{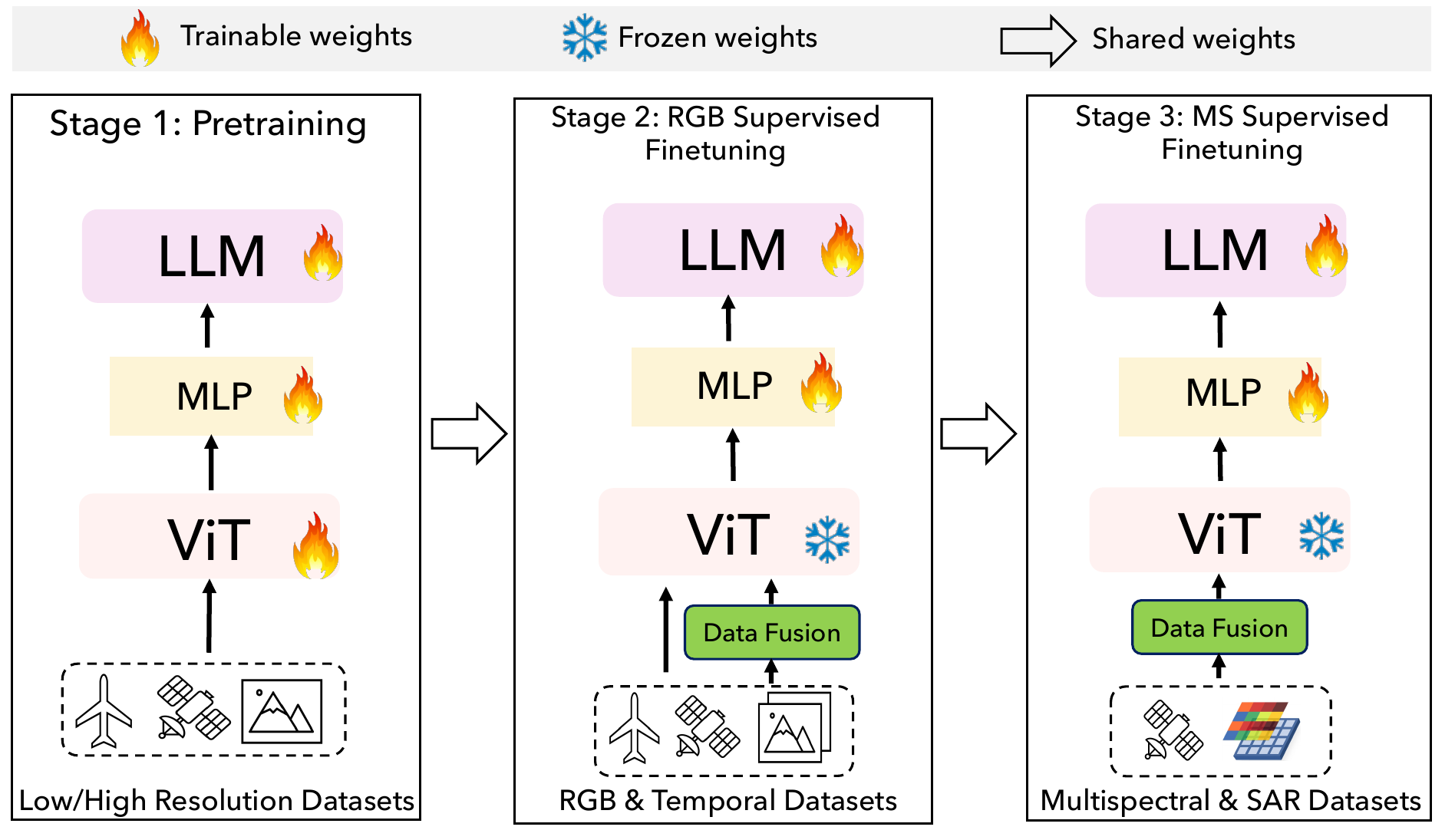}
\caption{EarthDial training Strategy for different RS modalities. We first pretrain with RGB imagery of different resolutions to achieve better alignment. Thereafter only LLM and projectors are trained on RGB and temporal inputs. We then expand the model's capability to multi-spectral and SAR imagery in Stage 3.}
 \label{fig:training_strategy}
 \vspace{-0.35cm}
\end{figure}

As illustrated in Fig.~\ref{fig:overall_arch}, EarthDial consists of three trainable components: a visual encoder, an MLP layer projector, and a large language model (LLM). 
Our model is relatively lightweight with only 4B parameters compared to the existing natural geospatial VLMs. 
The model is designed in such a way that it can take multi-resolution, multi-spectral, and time series datasets to generate various RS dialogues. 
As our visual encoder, we use InternViT-300M \cite{chen2024internvl}, a lightweight vision model distilled from the larger 6B InternViT that demonstrates strong visual encoding capability.  
Since our design goal is to have an efficient model, we use the Phi-3-mini pre-trained LLM \cite{abdin2024phi}.
To connect the visual encoder with the LLM, a simple MLP is used as the connector block to map visual tokens to the LLM space. 
As explained in Fig.~\ref{fig:training_strategy}, we tune the parameters of these three blocks systematically in different stages of training. 

Furthermore, the model incorporates two key modules, Adaptive High Resolution and Data Fusion, that are crucial in applying EarthDial to different resolution inputs as well as multi-spectral and multi-temporal RS data. 

\noindent\textbf{Adaptive High Resolution:} In remote sensing, images come in various sizes and resolutions, particularly high-resolution imagery where resizing for model can lead to the loss of critical pixel details. To address this, we adopt a dynamic resolution input strategy inspired by InternVL 1.5 \cite{chen2024far}, which enhances the model's ability to capture fine-grained details. The approach dynamically selects an optimal aspect ratio from a set of pre-defined ratios, dividing the image into 448×448-pixel tiles and creating a thumbnail for global context to help the model understand the overall scene.
Depending on the input resolution, 1 to 12 tiles can be created during training and upto 40 during inference. This approach minimizes aspect ratio distortion and accommodates varying resolutions during training and evaluation.

\noindent\textbf{Data Fusion:} The data fusion module in EarthDial is designed to improve the model’s capability to process multi-temporal, multi-spectral, synthetic aperture radar (SAR), and RGBI/hyper-spectral datasets. For multi-spectral inputs of any type, it operates by iteratively processing three channels of data at a time, which are passed through the Vision Transformer (ViT), to extract features for each channel. The extracted features are then aggregated and reduced in size using bilinear interpolation via the AnyRes block \cite{li2024llava} to ensure efficient handling of multi-spectral inputs. The AnyRes block splits the inputs into patches, encodes them, and uses bilinear interpolation to reduce tokens per patch, thus enabling the processing of multi-spectral inputs. These reduced visual embeddings are then concatenated with input corresponding text embeddings. The final step involves fusing these combined visual and textual features, which are passed to the LLM for further contextual processing. This fusion strategy allows EarthDial to integrate visual data from various modalities together with textual descriptions, improving its performance on complex RS tasks.  

For RGB temporal images, we first pass each image through the ViT to extract visual tokens, then stack and concatenate these tokens before passing the combined representation to the LLM.
Next, we explain our model training.

\subsection{Stage-wise Model Training}\label{sec:pretraining}
EarthDial is initialized with pretrained backbones explained in Sec.~\ref{sec:model_arch}. It is then trained in three distinct stages, designed to systematically expand model's capabilities.

\noindent\textbf{Stage 1 - RS Conversational Pretraining:}
The main goal of pretraining is to enhance the vision encoder’s ability to learn visual features from various satellite imagery sources, such as Sentinel-2, Landsat 8, Sentinel-1, and aerial images, and to generate descriptions for remotely sensed objects and scenes. During this phase, we use instruction datasets described earlier data section, specifically the Satlas and Skyscript datasets, which contain 7.6M image-text pairs for different types of remote sensing imagery. 

In the pretraining stage, the objective is to establish alignment between multiple sensor modalities in the RS domain with their corresponding natural language text descriptions.  
This is typically achieved using large datasets of image-text pairs, such as image and captions. 
The model learns to predict the corresponding text for a given image in an autoregressive manner, optimizing performance through a standard cross-entropy loss function. This process helps the model improve its ability to generate accurate text descriptions from remote sensing visual inputs.
As shown in Fig.~\ref{fig:training_strategy}, the architecture during pretraining is deliberately kept simple, only comprising of the vision encoder, MLP and the LLM, without any data fusion.  
At the pretraining stage, model is not introduced temporal and multispectral datasets, to keep the task relatively simple and initially learn strong representations from single-image RGB datasets from our proposed EarthDial-Instruct dataset.

At this stage, all the learnable components were trained to ensure proper alignment of the RS imagery. For pretraining the EarthDial model, we utilized 8 NVIDIA A100-80G GPUs. The training process employed an initial learning rate of 4e-5, optimized using a cosine learning rate scheduler. The key hyperparameters included the use of thumbnails to capture local features, an adaptive patch size ranging from 1 to 6 for capturing more detailed high-level features, a batch size of 2, and a maximum sequence length of 4096 tokens. Additionally, a weight decay of 0.01 was applied to regularize the model.

\noindent\textbf{Stage 2: RS RGB and Temporal finetuning:}\label{sec:finetuning1}
In the second stage of model training, we carefully selected high-quality instruction sets with prompts aimed at teaching the model to better understand user commands and successfully complete desired tasks on different satellite imagery. This approach allows Large Language Models (LLMs) to generalize to unseen tasks, improving zero-shot performance across a variety of remote sensing tasks. 

During this stage, we utilized previously pretrained model encoders and fine-tuned the MLP and LLM layers to perform visual instruction tuning for diverse remote sensing tasks, such as image captioning, classification, detection tasks (e.g., grounding, referring, identification), visual question answering (VQA), and temporal change detection. 

For multi-temporal images, we applied data fusion techniques, as explained in the model overview section, before passing the data to the feature extractor. Additionally, we used all the RGB and time-series datasets described in Tab.~\ref{tab:vqa_samples} during this phase of model fine-tuning. 

\noindent\textbf{Stage 3: RS Multispectral and SAR Finetuning:}\label{sec:finetuning2}
In the third stage of model training, we extended the model’s capabilities to work with multispectral, high-resolution RGBI, and SAR images. This was achieved by introducing a data fusion module, as explained earlier, to handle imagery with more than three or fewer than three spectral bands. The goal was to enable the model to learn from multispectral and SAR data for various remote sensing tasks. 

We utilized the pretrained weights from the previous stage 2 and fine-tuned the MLP and LLM layers, while keeping the ViT parameters frozen from Stage 1. The same ViT+MLP+LLM architecture was employed, with the addition of the data fusion module (Sec.~\ref{sec:model_arch}) to integrate information from multiple spectral channels. 
This training stage allows the model to handle a broader range of tasks, including land cover classification, species detection, Methane plume detection, UHI and SAR-based ship detection. Training hyperparameter details are given in Table~\ref{tab:hyperparams}.

\section{EarthDial-Instruct Dataset}
\label{sec:3ms}

\noindent\textbf{Pretraining Instruction Data:}
With EarthDial, our primary aim is to improve generalization performance on diverse downstream tasks, covering a wide range of modalities, multi-resolution, and multi-temporal data.
Therefore, we curate high-quality pre-train question-answer (QA) instruction pairs from SkyScript \cite{wang2024skyscript} and SatlasPretrain \cite{bastani2023satlaspretrain} data, which includes Sentinel-2 (S2), Sentinel-1 (SAR), NAIP, and Landsat imagery along with labels. We choose InternLM-XComposer2 \cite{dong2024internlm} for generating instructions using labels.  
Our data curation process involves filtering to ensure data quality. First, we filter out samples with sparse labels ($<$3).
Second, we apply luminance and coverage-based filtering to remove cloudy and low spatial coverage images.
Third, we prompt the  InternLM-XComposer2 to generate QA instruction pairs based on the key attributes (points, polygons, object category,  and position) specified in the inputs and labels. 
The details can be found in the supplementary material.
The curated instruction stats across different imagery sources are presented in Tab.~\ref{tab:vqa_samples}.

\begin{table}[t]
\centering

\resizebox{\linewidth}{!}{
\begin{tabular}{clll}
\toprule
\textbf{Stages} & \textbf{Datasets} & \textbf{Number of QA pairs} & \textbf{Token format} \\
\midrule

\multirow{4}{*}{\textbf{Stage 1}} & NAIP              & 3,000,113                  & [hr\_rgb\_0.5]    \\
                         & Sentinel-2        & 2,749,511                  & [s2\_rgb\_10]     \\
                         & Landsat           & 1,671,437                  & [l8\_rgb\_30]     \\
                         & SkyScript         & 249,855                    & [s2\_rgb\_10]     \\
\midrule
\multirow{9}{*}{\textbf{Stage 2}} & Classification           & 565,853     & [hr\_rgb\_0.5]             \\
                         & Detection               & 22,624       & [hr\_rgb\_0.5]             \\
                         & Visual Grounding        & 17,845       & [hr\_rgb\_0.5]             \\
                         & Caption                & 202,530     & [caption] [hr\_rgb\_0.5] \\
                         & VQA                    & 630,768     & [hr\_rgb\_0.5]             \\
                         & Change Detection       & 64,631      & [changedet][hr\_rgb\_temp\_0.5] \\
                         & Disaster assessment    & 37,563      & [hr\_rgb\_temp\_0.5] \\
                         & Geochat                & 308,861     & [hr\_rgb\_0.5] \\
\midrule
\multirow{6}{*}{\textbf{Stage 3}} & Sentinel -1             & 1,668,043   & [s1\_vh\_10]              \\
                         & Local Climate Zones & 765,591 & [s2\_ms\_30]              \\
                         & Tree Species           & 38,527       & [treeclassify] [hr\_rgbi\_0.5] \\
                         & Methane Plume          & 6,849        &  [hyper\_rgb\_3] \\
                         & Urban Heat Island      & 1,296        & [uhi][l8\_ms\_30]               \\
\bottomrule
\end{tabular}}
\vspace{-0.3cm}
\caption{Summary of the number of QA instruction pairs used during each stage, the image sources, and token formats.}
\label{tab:vqa_samples}
\end{table}

\begin{table}[t]
    \centering
    \setlength{\tabcolsep}{2pt}
     \resizebox{\linewidth}{!}{
        \begin{tabular}{cccccccccc}
            \toprule
            Stages & QA pairs & Epoch & Train-time & GPUs & Batchsize & Grad. Accu. & Lr & Decay & Adaptive patch sizes  \\
            \midrule
           2 & 1.8M & 1 & 4 (hours) & 4A100 80G & 2 &64 &$4e-5$ & 0.05 & 1 to 6\\
            3 & 2.4M & 1 & 6 (hours) & 4A100 80G& 2 &64 &$4e-5$ & 0.05 & 1 to 6\\
        \bottomrule
        \end{tabular}}
        \vspace{-2.5mm}
        \caption{Details of stages, hyperparams, \& training duration}
        \label{tab:hyperparams}
\end{table}

\begin{table*}[t]
    \centering\setlength{\tabcolsep}{3pt}
    \resizebox{\linewidth}{!}{
    \begin{tabular}{c | c c c c cc}
    \toprule
    \textbf{Model} & \textbf{AID~\cite{xia2017aid}} (RGB) & \textbf{UCMerced~\cite{yang2010bag}} (RGB) & \textbf{WHU-19~\cite{dai2010satellite}} (RGB) & \textbf{BigEarthNet~\cite{sumbul2019bigearthnet}} (RGB) & \textbf{xBD Set 1~\cite{gupta2019creating}} (Temporal) & \textbf{fMoW~\cite{christie2018functional}} (Temporal)\\
    \midrule

    GPT-4o & 74.73 & 88.76 & 91.14  & 49 & 67.95 & 21.43 \\
    InternVL-8B~\cite{chen2024internvl} & 60.4 & 58.23 & 79.3  & 19.73 & 51.44 & 21.04\\    
    GeoChat~\cite{kuckreja2024geochat} & 72.03 & 84.43 & 80.09  & 20.35 & 53.32 & 59.2\\

    \midrule
    \textbf{EarthDial} & \textbf{88.76} & \textbf{92.42} & \textbf{96.21}  & \textbf{68.82} & \textbf{96.37} & \textbf{70.03}\\
    
    \bottomrule
    \end{tabular}}
    \vspace{-0.3cm}
    \caption{Comparison of classification accuracy across various datasets. EarthDial indicates a significant improvement in classification accuracy over other existing generic and specialized VLMs.}
    \label{tab:classification}
\end{table*}

\begin{table*}[t]
    \centering
    
    \vspace{-0.3cm}\setlength{\tabcolsep}{5pt}
    \resizebox{\linewidth}{!}{
    \begin{tabular}{cccc|ccccc}
    \toprule
        \multirow{2}{*}{Method} & \multirow{2}{*}{BigEarthNet ~\cite{sumbul2019bigearthnet} (MS)} & \multirow{2}{*}{SoSAT-LCZ42~\cite{zhu2019so2sat} (MS)}  & \multirow{2}{*}{TreeSatAI~\cite{astruc2024omnisat} (RGBI)} & \multicolumn{5}{c}{Ship Dataset (SAR Imagery)} \\
        &  &  &  & Small & Medium & Large & Single & Multiple \\
        \midrule
        GPT-4o & 49 & 15.53 & 16.73 & 0.70 & 0.90 & 3.20 & 1.20 & 0 \\
        \textbf{EarthDial} & \textbf{69.94} & \textbf{60.72} & \textbf{56.61} & \textbf{12.14} & \textbf{26.02} & \textbf{35.56} & \textbf{26.03} & \textbf{6.06}\\
        
    \bottomrule
\end{tabular}}
\vspace{-0.3cm}
\caption{Performance evaluation of EarthDial across diverse modalities for multi-class classification and referred object detection tasks on SAR imagery. For MS modality, EarthDial achieves an average 32.5\% improvement in classification accuracy compared to GPT-4o. For the RGBI modality, EarthDial achieves 40.2\% higher accuracy than GPT-4o. A similar trend is observed for SAR imagery, where EarthDial delivers higher mAP@0.5, even when detecting multiple objects, highlighting the advantages of leveraging multi-modal inputs.}
\label{tab:ms_classification}
\end{table*}

\noindent\textbf{Downstream Tasks Image-text Instruction:}\label{subsec:downstream_tasks}
While pre-training focuses on enhancing generalization capabilities, we also need task-specific fine-tuning with diverse data types to improve downstream performance. 
To handle this, we carefully curate a large number of instruction-following datasets that cover ten diverse downstream tasks (e.g., scene classification, object detection, visual question answering, image captioning,  change detection, Methane plume detection, tree species classification, local climate zones, urban heat islands, and disaster assessment), six visual modalities (Optical, SAR, S2, Infrared, NIR, and Hyperspectral), and two visual temporal modalities (Optical and SAR).  
Details are in the supplementary.  

\noindent\emph{\textbf{Scene Classification:}}
We construct scene classification instructions with nine standard scene classifications, one multilabel scene classification (BigEarthNet \cite{sumbul2019bigearthnet}), and one temporal scene classification (FMoW \cite{christie2018functional}).  We limit the sequences to 4 images to handle multitemporal scene classification. 
We also use local climate zones (LCZ) \cite{zhu2019so2sat} and TreeSatAI-Time-Series \cite{astruc2024omnisat} datasets to determine the LCZs and botanical tree species, respectively.

\noindent\emph{\textbf{Object Detection:}}
We curate instruction-following tasks utilizing three tags i.e., \textit{refer, identify, and grounding} to perform region-level captioning, referring expressions, and grounded description for object detection datasets from various remote imaging modalities like optical, SAR, and infrared. %
We include visual grounding \cite{sun2022visual, zhan2023rsvg} datasets for region-level captioning.  Following  \cite{kuckreja2024geochat}, we compute the key attributes in the image such as 
the object's category, bounding box, color, relative position, and relative size 
present in the detection dataset. 
We present the box as $[x_{min}, y_{min}, x_{max}, y_{max}, \theta]$. Here, $(x_{min}, y_{min})$ denotes the top left corner point while  $(x_{max}, y_{max})$ presents the bottom right corner of the bounding box. 
The angle $\theta$ represents the rotation angle of the bounding box. 

\noindent\emph{\textbf{Visual Question Answering (VQA) \& Image Captioning:}} 
We create VQA and image captioning instructions by including six VQA and five image captioning datasets. %

\noindent\emph{\textbf{Change Detection:}} 
We integrate three binary change detection datasets and one multitemporal (MUDS \cite{yang2024made}) dataset. The original MUDS dataset has masks. Thus, to generate the instructions, we manually analyzed the images and masks and generated five captions for each sequence.

\noindent\emph{\textbf{Methane Plume Detection:}}
For the Methane plume detection, we utilize the STARCOP \cite{ruuvzivcka2023semantic} dataset which presents labeled hyperspectral (groundtruth mask and emission rate) data. %
We conversationally prompt three questions; (i) is there any Methane plume present in the input, (ii) what is the location of the plume, and (iii) what is its emission rate?

\noindent\emph{\textbf{Urban Heat Island (UHI):}}
We compute land surface temperature (LST) and normalized difference vegetation index (NDVI) maps from S2 and Landsat imagery.
Based on the data, we prompt to classify the underlying region into cooler, mildly hot, and extremely hot regions.  We also generate instructions about what underlying land use caused the temperature and how to mitigate it.

\noindent\emph{\textbf{Disaster Assessment:}}
We use the xBD \cite{gupta2019creating} dataset for building disaster assessment by utilizing bitemporal pre- and post-disaster images. 
We also include the QuakeSet \cite{cambrin2024quakeset} dataset and prompt to determine if an earthquake occurred between the bi-temporal SAR images and its magnitude.

\begin{table*}[t]
\centering
    \resizebox{\textwidth}{!}{
    \begin{tabular}{l|ccccc|ccccc|ccccc|ccccc}
    \toprule
    \multirow{2}{*}{Model} & \multicolumn{5}{c|}{\textbf{GeoChat-Instruct~\cite{kuckreja2024geochat}}} & \multicolumn{5}{c|}{\textbf{NWPU VHR-10~\cite{cheng2014multi}} (ZS)} & \multicolumn{5}{c|}{\textbf{Swimming Pool Dataset} (ZS)} & \multicolumn{5}{c}{\textbf{Urban Tree Crown Detection~\cite{zamboni2021benchmarking}} (ZS)} \\
    \cmidrule(lr){2-6} \cmidrule(lr){7-11} \cmidrule(lr){12-16} \cmidrule(lr){17-21}
     & Small & Medium & Large & Single & Multiple & Small & Medium & Large & Single & Multiple & Small & Medium & Large & Single & Multiple & Small & Medium & Large & Single & Multiple \\
    \midrule
    GeoChat~\cite{kuckreja2024geochat} & 2.9 & 13.6 & 21.7 & 16 & 4.3 & 2.5 & 3.2 & 14.7 & 13.23 & 1.9 & - & 3.1 & 7.3 & 1.2 & 0.6 & - & 1.8 & 8.9 & 2.9 & 3.1 \\
    InternVL2-4B~\cite{chen2024internvl} & 6.3 & 24.37 & 37.38 & 24.96 & 11.72 & 7.1 & 12.68 & 25.48 & 22.96 & 8.1 & 0.6 & 6.6 & 8.9 & 4.5 & 0.865 & -& 3.17 & 13.41 & 5.9 & 3.1 \\
    InternVL2-8B~\cite{chen2024internvl} & 7.20 & 23.76 & 31.99 & 25.77 & 9.30 & 4.26 &	11.85 & 20.72 & 21.66 & 5.86 & 0.3 & 4.7 & 18.27 & 7.6 & 0.514 & 0.6 & 3.99 & 17.1 & 7.9 & 3.94 \\
    	
    \midrule 
    \textbf{EarthDial} & \textbf{11.43} & \textbf{31.76} & \textbf{39.07} & \textbf{34.29} & \textbf{13.41} & \textbf{11.66} & \textbf{14.21} & \textbf{23.12} & \textbf{25.37} & \textbf{8.9} & \textbf{1.04} & \textbf{7.4} & \textbf{24.90} & \textbf{8.4} & \textbf{1.04} & \textbf{1.1} & \textbf{7.01} & \textbf{25.67} & \textbf{11.13} & \textbf{6.7}\\
    \bottomrule
    \end{tabular}
    }    
    \vspace{-0.3cm}
    \caption{Comparison of our EarthDial 
    for referred object detection tasks across various datasets. Small, medium, and large denote the object size, while single and multiple denote the number of objects. Here, ZS means zero-shot evaluation.}
    \label{tab:detection}
\end{table*}

\begin{table*}[t]
    \centering
    \resizebox{\textwidth}{!}{
    \begin{tabular}{lccc|ccc|ccc|ccc|ccc|ccc|ccc|ccc}
        \toprule
        \textbf{Model} & \multicolumn{3}{c|}{\textbf{GeoChat-Instruct~\cite{kuckreja2024geochat}}} & \multicolumn{3}{c|}{\textbf{HIT UAV~\cite{suo2023hit}}  (ZS)} & \multicolumn{3}{c|}{\textbf{NWPU VHR 10~\cite{NWPU_VHR_10}} (ZS)} & \multicolumn{3}{c|}{\textbf{SAR-Ship Dataset~\cite{wang2019sar}}} & \multicolumn{3}{c|}{\textbf{SRSDD-v1.0~\cite{lei2021srsdd}}} & \multicolumn{3}{c|}{\textbf{Swimming Pool} (ZS)} & \multicolumn{3}{c|}{\textbf{UCAS AOD~\cite{zhu2015orientation}} (ZS)} & \multicolumn{3}{c}{\textbf{Urban Tree Crown~\cite{zamboni2021benchmarking}} (ZS)} \\
        ZS=Zero-Shot & R-1 & R-L & MT & R-1 & R-L & MT & R-1 & R-L & MT & R-1 & R-L & MT & R-1 & R-L & MT & R-1 & R-L & MT & R-1 & R-L & MT & R-1 & R-L & MT \\
        \midrule
        GPT-4o & 9.41 & 7.6 & 8.02 & 10.96 & 9.02 & 8.23 & 17.68 & 11.81 & 9.63 & 7.49 & 7.24 & 7.07 & 6.9 & 6.67 & 7.94 & 13.94 & 10.19 & 7.91 & - & - & - & 11.63 & 10.11 & 7.12 \\
        InternVL2-8B~\cite{chen2024internvl} & 10.58 & 9.06 & 8.5 & 11 & 9.53 & 8.4 & 11.88 & 9.63 & 7.7 & 9.67 & 8.67 & 8.19 & 10.55 & 8.84 & 8.94 & 14.63 & 12 & 6.95 & 14.52 & 10.43 & 8.59 & 11.89 & 9.8 & 6.79 \\
        GeoChat~\cite{kuckreja2024geochat} & 72.77 & 72.74 & 61.9 & 59.85 & 59.85 & 51.31 & 65.02 & 65.02 & 53.31 & 57.15 & 57.15 & 52.2 & 63.72 & 63.72 & 57.31 & \textbf{64.73} & \textbf{64.73} & \textbf{51.47} & \textbf{65.03} & \textbf{65.03} & 52.4 & 60.52 & 60.52 & 50.48 \\     \midrule
        \textbf{EarthDial} & \textbf{73.38} & \textbf{73.34} & \textbf{62.72} & \textbf{61.83} & \textbf{61.83} & \textbf{52.80} & \textbf{72.14} & \textbf{72.14} & \textbf{60.01} & \textbf{63.1} & \textbf{63.1} & \textbf{54.83} & \textbf{68.8} & \textbf{68.8} & \textbf{62.45} & 61.96 & 61.96 & 47.42 & 64.03 & 64.03 & \textbf{52.82} & \textbf{63.47} & \textbf{63.47} & \textbf{54.09} \\
        
        \bottomrule
    \end{tabular}%
    }
    \vspace{-0.3cm}
    \caption{Comparison of our EarthDial with existing generic and specialized VLMs for region captioning task across various datasets.}
    
    \label{tab:region_captioning}
\end{table*}

\begin{table*}[!t]
    \centering
    \resizebox{\textwidth}{!}{
    \begin{tabular}{l|ccccc|ccccc|ccccc|ccccc}
        \toprule
        \multirow{1}{*}{\textbf{Model}} & \multicolumn{5}{c|}{\textbf{HIT UAV~\cite{suo2023hit}} (ZS)} & \multicolumn{5}{c|}{\textbf{NWPU VHR 10~\cite{cheng2014multi}} (ZS)} & \multicolumn{5}{c|}{\textbf{Swimming Pool Dataset} (ZS)} & \multicolumn{5}{c}{\textbf{UCAS AOD~\cite{zhu2015orientation}} (Zero-Shot)} \\ 
        & \textbf{\texttt{@}0.5} & \textbf{\texttt{@}0.25} & \textbf{R-1} & \textbf{R-L} & \textbf{MT} & \textbf{\texttt{@}0.5} & \textbf{\texttt{@}0.25} & \textbf{R-1} & \textbf{R-L} & \textbf{MT} & \textbf{\texttt{@}0.5} & \textbf{\texttt{@}0.25} & \textbf{R-1} & \textbf{R-L} & \textbf{MT} & \textbf{\texttt{@}0.5} & \textbf{\texttt{@}0.25} & \textbf{R-1} & \textbf{R-L} & \textbf{MT} \\
        \midrule
        GPT-4o & 0.1 & 0.7 & 14.20 & 10.56 & 7.16 & 0.7 & 6.1 & 14.72 & 10.82 & 9.41 & 0.1 & 1.2 & 12.87 & 10.07 & 7.79 & 0.1 & 1.3 & 14.71 & 11.14 & 5.97 \\ 
        InternVL2-4B~\cite{chen2024internvl} & 0.6 & 6.4 & 28.1 & 27.68 & 23.94 & 10.6 & 29.87 & \textbf{30.67} & \textbf{29.09} & 21.92 & 0.8 & 4.2 & 28.3 & 28.08 & \textbf{24.64} & 4.6 & 31.8 & 21.01 & 20.01 & 11.65\\ 
        GeoChat~\cite{kuckreja2024geochat} & 0.8 & 8.0 & 22.82 & 22.22 & \textbf{22.27} & 2.2 & 15.27 & 21.46 & 20.74 & 21.38 & 1.8 & \textbf{8.8} & 21.45 & 21.15 & 23.94 & 1.45 & 13.63 & 20.02 & 18.81 & \textbf{14.22} \\ 
        \midrule
        \textbf{EarthDial} & \textbf{2.61} & \textbf{13.86} & \textbf{28.31} & \textbf{28.06} & 22.25 & \textbf{17.07} & \textbf{41.00} & 27.05 & 26.35 & \textbf{23.12} & \textbf{1.9} & 7.4 & \textbf{29.7} & \textbf{29.31} & 22.77 & \textbf{8.5} & \textbf{34.02} & \textbf{21.17} & \textbf{20.28} & 13.01\\
        \bottomrule
    \end{tabular}%
    }    
    \vspace{-0.3cm}
    \caption{Comparison of EarthDial with existing generic and specialized VLMs on the grounding description task across multiple datasets.}
    \label{tab:grounding_description}
\end{table*}

\begin{table*}[!t]
    \centering
    \resizebox{\textwidth}{!}{
    \begin{tabular}{lccccccccccccccc}
        \toprule
        \multirow{2}{*}{\textbf{Model}} & \multicolumn{3}{c}{\textbf{NWPU RESISC45 Captions~\cite{cheng2017remote}}} & \multicolumn{3}{c}{\textbf{RSCID Captions~\cite{lu2017exploring}}} & \multicolumn{3}{c}{\textbf{RSITMD Captions~\cite{yuan2022exploring}} (Zero-shot eval)} & \multicolumn{3}{c}{\textbf{Sydney Captions~\cite{qu2016deep}}} & \multicolumn{3}{c}{\textbf{UCM Captions~\cite{qu2016deep}}} \\
        \cmidrule(lr){2-4} \cmidrule(lr){5-7} \cmidrule(lr){8-10} \cmidrule(lr){11-13} \cmidrule(lr){14-16}
         & \textbf{Rouge1} & \textbf{Rouge-L} & \textbf{Meteor} & \textbf{Rouge1} & \textbf{Rouge-L} & \textbf{Meteor} & \textbf{Rouge1} & \textbf{Rouge-L} & \textbf{Meteor} & \textbf{Rouge1} & \textbf{Rouge-L} & \textbf{Meteor} & \textbf{Rouge1} & \textbf{Rouge-L} & \textbf{Meteor} \\
        \midrule
        GPT-4o & 19.43 & 14.86 & 28.16 & 20.53 & 15.59 & 26.03 & 18.31 & 14.22 & 24.83 & 14.52 & 12.55 & 18.87 & 25.77 & 20.58 & 33.18 \\
        InternVL2-8B~\cite{chen2024internvl} & 20.69 & 15.64 & 30.18 & 21.59 & 16.13 & 28.17 & 18.91 & 14.65 & 26.02 & 15.71 & 13.8 & 19.69 & 22.9 & 17.91 & 28.61 \\
        GeoChat~\cite{kuckreja2024geochat} & 14.86 & 12.54 & 15.21 & 13.48 & 11.59 & 12.39 & 13.41 & 11.5 & 12.33 & 12.0 & 11.26 & 10.63 & 14.4 & 13.22 & 14.27 \\
        \midrule
        \textbf{EarthDial} & \textbf{45.84} & \textbf{39.96} & \textbf{80.61} & \textbf{33.77} & \textbf{27.61} & \textbf{56.18} & \textbf{26.74} & \textbf{21.72} & \textbf{34.06} & \textbf{49.39} & \textbf{41.0} & \textbf{57.31} & \textbf{40.0} & \textbf{34.15} & \textbf{51.42} \\
        
        \bottomrule
    \end{tabular}
    }
    \vspace{-0.3cm}
    \caption{Comparison of our EarthDial with existing generic and specialized VLMs for Image captioning tasks across various datasets.}
    
    \label{tab:image_captioning}
\end{table*}
\begin{table}[!t]
    \centering\setlength{\tabcolsep}{3pt}
    \resizebox{\linewidth}{!}{
    \begin{tabular}{lcccc|lccc}
    \toprule
    \textbf{Model} & \textbf{Presence} & \textbf{Comp} & \textbf{R/U} & \textbf{Avg.} & \textbf{Model} & \textbf{Presence} & \textbf{Comp} & \textbf{Avg.} \\
    \midrule
    MiniGPTv2 & 55.16 & 55.22 & 39.00 & 54.96 & MiniGPTv2 & 40.79 & 50.91 & 46.46 \\
    Qwen-VL~\cite{bai2023qwen} & 38.57 & 67.59 & 61.00 & 55.35 & Qwen-VL~\cite{bai2023qwen} & 66.44 & 60.41 & 63.06 \\ 
    InternVL2-8B~\cite{chen2024internvl} & 58.54 & 72.28 & 71.00 & 66.51 & InternVL2-8B~\cite{chen2024internvl} & \textbf{67.35} & 76.91 & \textbf{72.70} \\
    GeoChat~\cite{kuckreja2024geochat} & 91.09 & 90.33 & 94.00 & 90.70 & GeoChat~\cite{kuckreja2024geochat} & 58.45 & \textbf{83.19} & 72.30 \\
    
    LHRS-Bot~\cite{muhtar2024lhrs} & 88.51 & 90.00 & 89.07 & 89.19 & EarthGPT~\cite{zhang2024earthgpt} & 62.77 & 79.53 & 72.06 \\
    
    \midrule
    \textbf{EarthDial} & \textbf{92.58} & \textbf{92.75} & \textbf{94} & \textbf{92.70} & \textbf{EarthDial} & 58.89 & 83.11 & 72.45 \\
    
    \bottomrule
    \end{tabular}}
    \vspace{-0.3cm}
    \caption{Comparison of EarthDial with existing VLMs for visual question answering task (left: RSVQA-LRBEN, right: RSVQA-HRBEN). Comp: Comparison, R/U: Rural/Urban.}
    \label{tab:vqa}
\end{table}

\section{Experiments}
\label{sec:experiments}
    Here, we discuss the experimental results of the proposed EarthDial across a diverse set of applications, including RGB, multispectral, SAR, infrared, and thermal imagery. Our evaluation covers various tasks such as scene classification, referred object detection, region captioning, grounding descriptions, VQA, image captioning, change detection, and methane plume detection.\newline
    \noindent\textbf{Scene classification:}
        For zero-shot evaluation, we compare EarthDial with RGB datasets in Tab. \ref{tab:classification}, whereas BigEarthNet (RGB), xBD Set 1, and fMoW are supervised. We also compare multi-spectral (MS) datasets (BigEarthNet, SoSAT-LCZ42), and the RGB-Infrared TreeSatAI dataset as shown in Tab. \ref{tab:ms_classification}. We notice that EarthDial shows consistent performance gain against existing generic and specialized VLMs. Moreover, EarthDial outperforms temporal scene classification FMoW as well as over xBD test-set 1 for disaster assessment \cite{gupta2019creating} as in Tab. \ref{tab:classification}. Further results are in a supplementary document.
      
    \noindent\textbf{Object Detection:}
        Following~\cite{kuckreja2024geochat}, we address three sub-tasks: referred object detection, region captioning, and grounding description. We consider existing generic VLMs like GPT-4o, InternVL2-4B while specialized GeoChat for the comparison. As existing InternVL2 doesn't provide the rotated bounding boxes, for fair comparison, we finetune the InternVL2 on GeoChat-Instruct and compared it with our EarthDial. Table~\ref{tab:detection}, ~\ref{tab:region_captioning}, and  ~\ref{tab:grounding_description} depict 
        our EarthDial is a clear winner and consistently outperforms the all other compared VLMs by a large margin. Especially on grounding description tasks, where existing methods struggle to detect/localize the objects, our Earthdial achieves higher mAP. Also, improved results on SAR imagery datasets showcase the multi-modal data processing capability of our EarthDial.

    \noindent\textbf{Image Captioning and Visual Question Answering: }
        Our EarthDial outperforms the existing generic and specialized VLMs by clear margins over image captioning datasets as shown in Tab. \ref{tab:image_captioning}. In addition, for the VQA task, we utilize datasets RSVQA-LRBEN and RSVQA-HRBEN (zero-shot) for evaluation, following~\cite{kuckreja2024geochat}. 
        Tab.~\ref{tab:vqa} presents the VQA accuracy of existing models compared to the proposed EarthDial, outperforming most of the categories.

\begin{table}[t]
    \centering
    \setlength{\tabcolsep}{3pt}
    \resizebox{\linewidth}{!}{
    \begin{tabular}{lcccccccccccc}
    \toprule
    \multirow{2}{*}{\textbf{Model}} & \multicolumn{3}{c}{\textbf{Dubai CC}} & \multicolumn{3}{c}{\textbf{LEVIR MCI~\cite{liu2024change}}} & \multicolumn{3}{c}{\textbf{MUDS~\cite{yang2024made}}} & \multicolumn{3}{c}{\textbf{SYSU~\cite{noman2024cdchat} (zero-shot)}}\\
    \cmidrule(lr){2-4} \cmidrule(lr){5-7} \cmidrule(lr){8-10} \cmidrule(lr){11-13}
    & R-1 & R-L & MT & R-1 & R-L & MT & R-1 & R-L & MT & R-1 & R-L & MT \\
    \midrule
    GPT-4o & 8.81 & 7.45 & 18.68 & 10.33 & 8.4 & 22.05 & 14.18 & 11.02 & 20.92 & 16.48 & 12.32 & \textbf{17.49} \\
    InternVL2-4B~\cite{chen2024internvl} & 7.31 & 6.38 & 21.12 & 8.88 & 7.43 & 22.14 & 10.25 & 7.9 & 17.73 & 13.27 & 9.98 & 14.36 \\
    GeoChat~\cite{kuckreja2024geochat} & 14.21 & 14.19 & 28.91 & 17.15 & \textbf{35.42} & 12.35 & 12.28 & 12.23 & 15.98 & 13.45 & 12.02 & 13.96\\
    \midrule
    \textbf{EarthDial} & \textbf{31.94} & \textbf{30.66} & \textbf{55.83} & \textbf{33.78} & {30.47} & \textbf{74.8} & \textbf{28.16} & \textbf{24.03} & \textbf{33.56} & \textbf{18.03} & \textbf{17.42} & 14.98 \\
   
    \bottomrule
    \end{tabular}
    }
    \vspace{-0.3cm}
    \caption{Comparison of EarthDial with existing generic and specialized VLMs on the change detection task.}
    \label{tab:change_Detection}
\end{table}
\begin{table*}[t]
    \centering
    \resizebox{\linewidth}{!}{
        \begin{tabular}{l|ccc|cc|ccc|cc|cc}
            \toprule
            \multirow{2}{*}{\textbf{Model}} & \multicolumn{3}{c|}{\textbf{Image Captioning}} & \multicolumn{2}{c|}{\textbf{Region Classification}} & \multicolumn{3}{c|}{\textbf{Image Classification}} & \multicolumn{2}{c|}{\textbf{Object Detection}} & \multicolumn{2}{c}{\textbf{Referred Object Detection}} \\
            & R-1 & R-L & MT & Test Set-1 & Test Set-2 & Test Set-1 & Test Set-2 & Test Set-3 & mAP@0.5 & mAP@0.25 & mAP@0.5 & mAP@0.25 \\
            \midrule
            GPT-4o	& 14.21	& 10.35 & 19.52 & 51.68 & 71.62 & 67.95 & 75.45 & \textbf{70.41} & 0.2 & 2.15 & 0 & 0 \\
            InternVL2-8B & 13.89 & 10.37 & 14.92 & 14.39 & 58.33 & 51.44 & 61.52 & 51.12 & 0.6 & 1.07 & 0 & 0.7\\
            GeoChat & 14.18 & 10.67 & 12.20 & 25.30 & 57.65 & 53.32 & 52.19 & 49.51 & 1.15 & 7.2 & 0.2 & 3.09\\
            \midrule
            \textbf{EarthDial} & \textbf{87.26} & \textbf{87.26} & \textbf{88.53} & \textbf{53.7} & \textbf{83.09} & \textbf{96.37} & \textbf{82.85} & 54.01 & \textbf{7.6} & \textbf{21.11} & \textbf{5.1} & \textbf{13.09} \\
            \bottomrule
        \end{tabular}}
    \vspace{-0.3cm}
    \caption{Comparison of our EarthDial for various tasks on the xBD dataset (temporal). R-1, R-L, and MT denote ROUGE-1, ROUGE-L, and METEOR scores, respectively. \textbf{Image Captioning:} Describes the damage observed in the post-disaster image. \textbf{Region Classification:} Test Set-1, model classifies the level of damage in the user-marked region. Test Set-2, binary classification of user marked region into 'damage' or 'no-damage' class. \textbf{Image Classification:} Test Set-1, classifies the image into the type of disaster. Test Set-2, binary classification of image into 'damage' or 'no-damage' class. Test Set-3, classifies the image based on affected building count (none, one, two, or many). Classification recall is used for the evaluation of Region and Image Classification tasks. \textbf{Object Detection:} Locates all large buildings in the post-disaster image. \textbf{Referred Object Detection:} Locates the user-referred damage region/object in the post-disaster image. It is clearly observed that our EarthDial significantly improves performance for all the tasks. }
    \label{table:xBD_eval}
\end{table*} 

    \noindent\textbf{Change Detection:}
        To demonstrate the temporal data processing capability of the proposed EarthDial, we evaluate its performance on a change detection task. 
        We applied the data fusion strategy (discussed in  Sec.~\ref{sec:model_arch}) to merge tokens obtained from multiple images across time.
        The results in Tab. \ref{tab:change_Detection} highlight EarthDial's strong ability to interpret and respond effectively to temporal data.\newline
    \noindent\textbf{Temporal Disaster Assesment:}
        Here, we show the capability of our EarthDial in processing temporal data. First, we consider the benchmark xBD dataset belonging to the disaster assessment task for the experiment. xBD dataset has two images: pre-disaster and post-disaster. Thus, from xBD dataset, we prepare eight sub-tasks covering temporal {image captioning, region classification, image classification, object detection, and referred object detection} applications.
        Briefly described as below:\newline
        \textit{Image Captioning:} Compared to the pre-disaster image, describe the damage observed in the post-disaster image.\newline
        \textit{Region Classification:} We have two sets here. In Test Set-1, classification of the level of damage in the user-marked region is included. In Test Set-2, binary classification of the user-marked region into 'damage' or 'no-damage' class is added.\newline
        \textit{Image Classification:} We have three sets here. In test Set-1, the classification of the image into the type of disaster (volcano, fire, earthquake, flood, tsunami, wind) is included. In Test Set-2, binary image classification into the 'damage' or 'no-damage' class is added. In Test Set-3, examples classify the image based on the affected building count (none, one, two, or many).\newline
        \textit{Object Detection:} Locates all large buildings in the post-disaster image.\newline
        \textit{Referred Object Detection:} Locates the user-referred damage region/object in the post-disaster image.\newline
        In the main paper, we report results for \textit{Image Classification test Set-1}. Here, Tab.~\ref{table:xBD_eval} presents a summary of the performance results of both generic and specialized VLMs across the sub-tasks discussed above. EarthDial consistently outperforms all other existing VLMs by a significant margin, demonstrating its capability to effectively process temporal data for the desired task.
        Tab.~\ref{table:xBD_eval} presents a summary of the performance results of both generic and specialized VLMs across the sub-tasks discussed above. EarthDial consistently outperforms all other existing VLMs by a significant margin, demonstrating its capability to effectively process temporal data for the desired task.\footnote{R-1, R-L, MT denotes ROUGE-1, ROUGE-L, METEOR scores}
        Additionally, we evaluated our method on QuakeSet \cite{cambrin2024quakeset} for earthquake prediction using SAR imagery. We evaluate the model based on binary classification to determine whether an earthquake event occurred or not between the input SAR imagery. While GPT-4o achieves a classification accuracy of 55.86, our method outperforms it with an accuracy of 57.53.\newline
    \noindent\textbf{Multi-modal Data Processing:}
        To showcase the capability of EarthDial in processing multi-modal data, we examine multi-spectral (MS), RGB-infrared, and SAR imagery for classification and referred object detection tasks. Comparative results of the proposed EarthDial and existing GPT-4o are given in Tab.~\ref{tab:ms_classification}. Our EarthDial outperforms GPT-4o by a significant margin on both multi-spectral, RGBI, and SAR imagery. Significant improvement in the performance highlights the effectiveness of our multi-band fusion strategy.\newline
    \noindent\textbf{Urban Heat Island:}
        For UHI, we prompt to classify the underlying region into cooler, mildly hot, and extremely hot regions.
        We achieve an accuracy of 56.77\% in identifying the temperature trends from user-input Landsat8 bands, whereas GPT-4o achieves an accuracy of 22.68\% in the same task. This shows the capability of our EarthDial in processing multi-modality data effectively compared to the existing generic GPT-4o.\newline
    \noindent\textbf{Methane Plume Classification:}
        We consider STARCOP dataset for the evaluation, which has a four-channel image (RGB + mag1c band). 
        Our EarthDial processes these RGBM bands to predict the presence of methane plumes as "Yes" or "No".  We compare our accuracy with the GPT4o model. We simply give RGB and Mag1c bands to the GPT4o and collect the response. With this, GPT4o achieves an accuracy of 40.93\%, while our EarDial achieves 77.09\% \textit{i.e.,} improves it by 32.16\%.

\begin{figure*}[t]
\centering
 \includegraphics[width=1\linewidth]{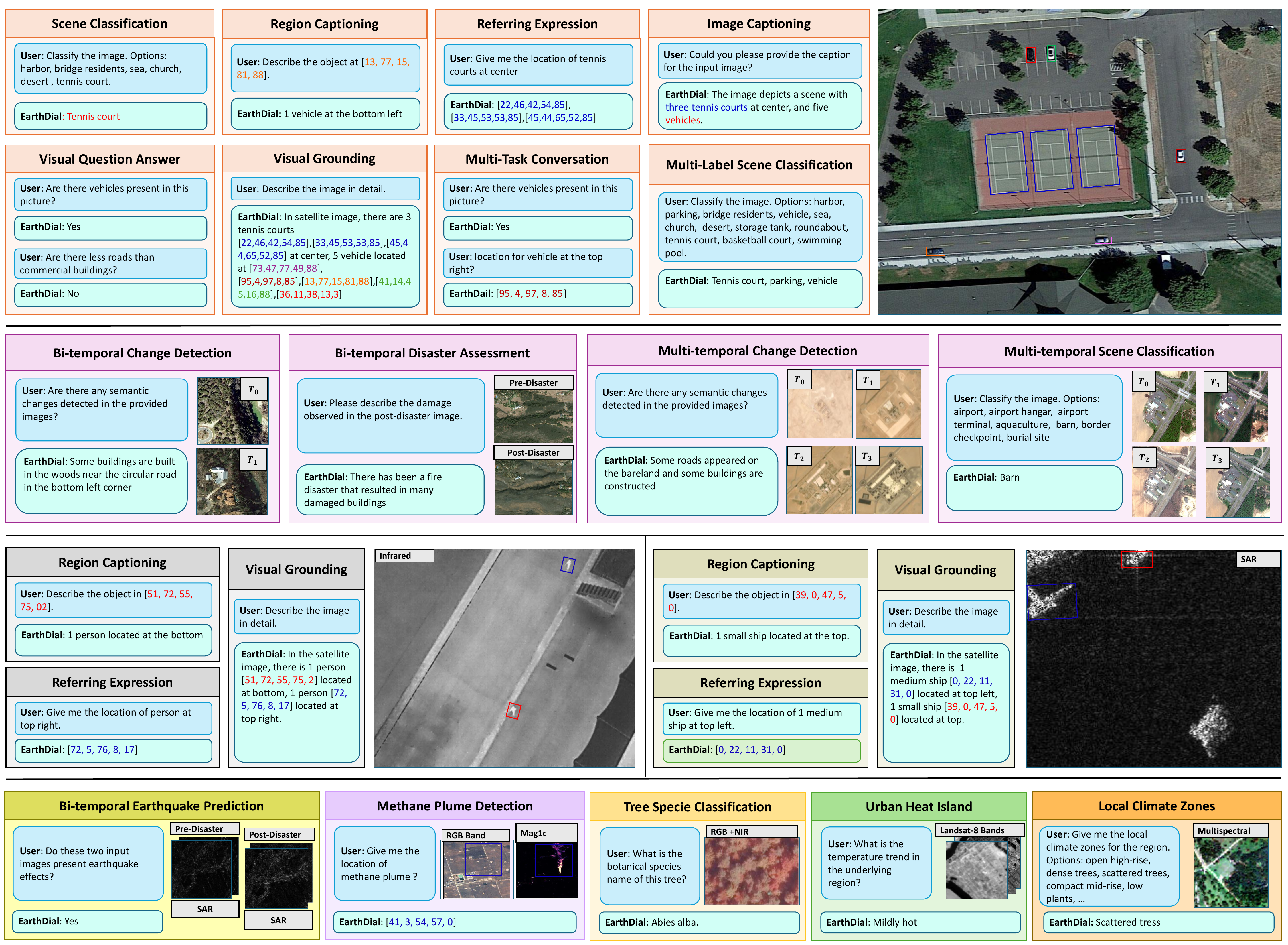} 
\caption{Illustration of our versatile \textbf{EarthDial} model that performs across multi-modalities, multi-resolution, multispectral, and multitemporal data from diverse remote sensing applications. \textbf{EarthDial} extends its capabilities to a range of tasks such as scene classification, image/region-captioning, referring expression, VQA, referring expression, object detection, temporal change/disaster detection, Methane plume detection, tree species classification, UHI, and LCZs detection across multi-modalities, multi-resolution remote sensing data.}
 \label{fig:Downstream_tasks_qa_pairs}
\end{figure*}

\section{Ablation Study}
Here, we demonstrate the effectiveness of our multi-stage pre-training, and multi-spectral band fusion strategy.

\noindent \textbf{Effect of Multi-Stage Pre-Training:} We assess the performance of the EarthDial model on a complex detection task, both before and after multi-stage pre-training. For this analysis, we focus on the referred object detection task using the Geochat-Instruct dataset. With multi-stage pre-training, EarthDial’s mAP @0.5 improves by 5\%, showing a notable boost in detecting multiple referred objects compared to its performance without pre-training.

\begin{table}[t]
\centering
\setlength{\tabcolsep}{3pt}
\resizebox{\linewidth}{!}{
\begin{tabular}{ccccccc|ccc}
\toprule
\multicolumn{7}{c|}{Referred object detection task (GeoChat-Instruct dataset~\cite{kuckreja2024geochat})} & \multicolumn{3}{c}{Multi-band classification} \\

\multirow{2}{*}{AHR} & \multirow{2}{*}{Pre-training} & \multirow{2}{*}{Small} & \multirow{2}{*}{Medium} & \multirow{2}{*}{Large} & \multirow{2}{*}{Single} & \multirow{2}{*}{Multiple} & Data & BigEarthNet &  TreeSat\\
& &  &  &  &  &  & Fusion & MS &  AI \\
\midrule
\ding{55} & \ding{55} & 6.06 & 22.97 & 36.26 & 24.75 & 10.3 & Average & 47.66 & 49.09\\
\ding{55} & \ding{51} & 10.85 & 32.05 & 36.15 & 33.73 & 11.91 & Max-pooling & 34.62 & 29.84\\
\ding{51} & \ding{51} & 11.75 & 33.33 & 42.60 & 34.78 & 15.03 & Bilinear & 67.01 & 56.93\\
\bottomrule
\end{tabular}}
\vspace{-0.3cm}
\caption{Ablation study about the effect of multi-stage training (sec. \ref{sec:pretraining}) and spectral band fusion strategy (sec. \ref{sec:model_arch}). The proposed multi-stage pre-training approach improves EarthDial's mAP @0.5 for detecting multiple referred objects by 5\%. Additionally, the use of our bilinear fusion strategy enhances the average classification accuracy by 9.5\%.}
\label{tab:ablation}
\end{table}

\noindent \textbf{Effect of Multispectral Band Fusion:} In this setup, we evaluate EarthDial's performance with respect to the data fusion module (using average pooling and bilinear interpolation) for multi-spectral band fusion, on the classification tasks for BigEarthNet and TreeSat AI. Table~\ref{tab:ablation} shows that the bilinear interpolation fusion strategy is more effective than simple average pooling, enhancing model's average accuracy by 13.5\%. The impact of multi-spectral data combined with our multi-band fusion strategy is evident in Table~\ref{tab:ms_classification}. EarthDial achieves a 1.75\% improvement in classification accuracy on the Multi-Spectral (MS) version of BigEarthNet compared to its RGB counterpart. This demonstrates EarthDial's ability to leverage complementary information from multispectral bands, enhancing performance in multi-class classification tasks.

\section{Conclusion}
\label{sec:conc}
    We present EarthDial, a conversational assistant purpose-built for Earth Observation (EO) data, capable of transforming complex, multi-sensory Earth observations into interactive, and natural language dialogues. EarthDial supports multi-spectral, multi-temporal, and multi-resolution imagery as input and addresses a wide spectrum of remote sensing tasks, including classification, detection, captioning, question answering, visual reasoning, and visual grounding.
    To enable this versatility, we developed a comprehensive instruction-tuning dataset containing over 11M instruction pairs, encompassing diverse modalities such as RGB, S2, Synthetic Aperture Radar (SAR), Near-Infrared (NIR), and infrared. Additionally, EarthDial excels in handling bi-temporal and multi-temporal sequence analysis, making it highly effective for applications like change detection/disaster assessment.
    Our experiments across 44 downstream tasks highlight EarthDial's superior performance over existing generic and domain-specific models, demonstrating its robust generalization capabilities and its potential to set a new standard in EO task automation.

\appendix
\counterwithin{figure}{section}
\counterwithin{table}{section}

\section*{Appendix}
    Here, we first provide details about the EarthDial-Instruct dataset used to train our model, in three stages. 
    Second, we conduct an ablation study comparing the performance of the EarthDial model fine-tuned with LoRA against the fully fine-tuned version, evaluating both models on zero-shot detection datasets.
    Last, we provide more qualitative analysis of our EarthDial model, compared to recent state-of-the-art VLMs, demonstrating its better generalization across multi-modalities, multi-resolution, and multi-temporal downstream EO tasks.

\section{EarthDial-Instruct Dataset}
\label{sec:supp_3ms}

The fundamental objective of constructing domain-specific VLM is to improve generalization performance on diverse downstream tasks, covering a wide range of modalities, multi-resolution, and multi-temporal data. 
Therefore, we curate high-quality pre-train question-answer (QA) instruction pairs from SkyScript \cite{wang2024skyscript} and SatlasPretrain \cite{bastani2023satlaspretrain} data, which includes Sentinel-2 (S2), Sentinel-1 (SAR), NAIP, and Landsat imagery along with labels. 
Specifically, we choose InternLM-XComposer2 \cite{dong2024internlm} as an instruction generator after evaluating its generation outputs against state-of-the-art leading VLMs at the time of selection, where it demonstrated superior efficiency in handling large-scale data for generating vision QA instruction pairs. 
The methodology involved multiple steps of filtering to ensure the quality of the data, as depicted in \cref{fig:vqa-generation-pipeline}. In step I we proceed with a label-based filtering, where we filter out samples that are associated with at least three labels, ensuring that each image contained enough descriptive content to support meaningful instruction samples. In step II, an image-based filtering is applied, where we apply luminance and coverage-based filtering to remove cloudy images as well as low spatial coverage images. More specifically, we apply a threshold on the average luminance and remove images with insufficient coverage.
In step III, we prompt the InternLM-XComposer2 to generate QA instruction pairs based on the key attributes (points, polygons, object category, and position) specified in the inputs and labels. These attributes, before being input in the processing pipeline, undergo formatting to natural language to be understood by the VLM. When processing a sample, we prompt the model multiple times, asking for a QA instruction set for each attribute specifically. Each prompt also contains information about all the other attributes detected in the image. Furthermore, in the same prompt, we provide an example of a satisfactory QA instruction set, sampled from a list of predefined instruction sets. The generation is repeated up to 5 times, if the expected format is not respected. We present the workflow explicitly below:

\begin{figure}[t]
    \centering
    \includegraphics[width=\linewidth]{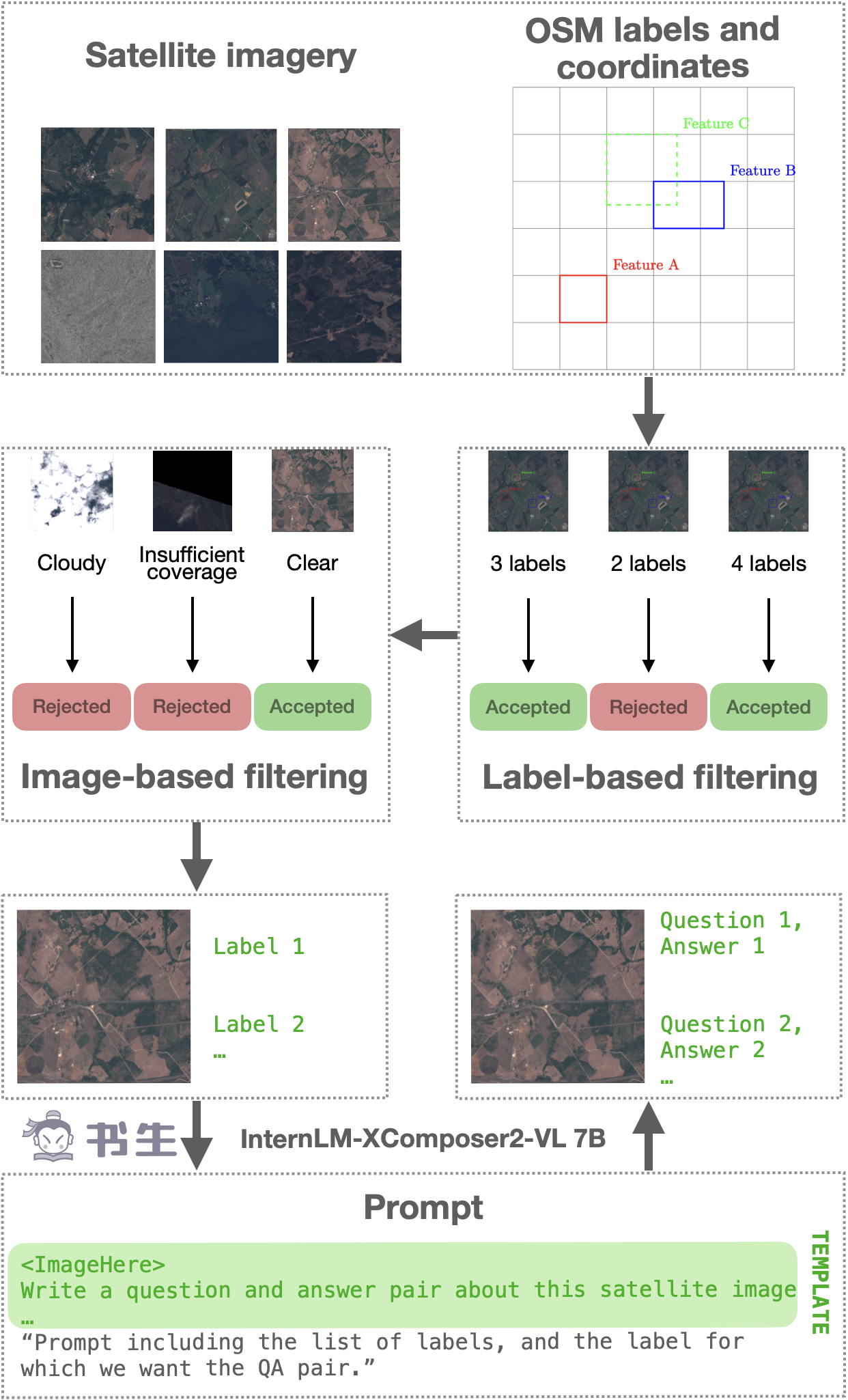}
    \caption{Overview of the data preparation and filtering pipeline used in the QA instruction dataset generation. The process begins with the pairing of OpenStreetMap (OSM) labels and their corresponding different sources of satellite imagery. The data goes through a label-based filtering process selecting only images with 3 labels or above, and then this data undergoes a second filtering process which is image-based to remove low-quality images. The high quality images remaining are then passed to the InternLM-XComposer2-VL model to generate question-answer pairs based on the associated reliable labels from OSM.}
    \label{fig:vqa-generation-pipeline}
\end{figure}

\begin{enumerate}
    \item A satisfactory QA instruction set example: \textit{Subject: parking lot. Question: How does the parking lot contribute to environmental sustainability? Answer: The parking lot in the lower left seems to be equipped with solar panel canopies, promoting renewable energy use.}
    \item The prompt: \textit{Write a question and answer pair about this satellite image. For example, on another image, a satisfactory pair is: \textcolor{blue}{satisfactory\_qa\_instruction}. The current image has been annotated with the following keywords: \textcolor{blue}{attribute\_1, attribute\_2, \dots}. Generate the pair for the following subject: \textcolor{blue}{attribute\_1}, which is visible in the satellite image. The question or answer must refer to the \textcolor{blue}{attribute\_1}, and must refer to either its position, interaction with other elements in the image, characteristics, or function. The answer must be objective, based on visible elements in the image, and require the image to answer. Avoid any assumptions or extrapolations that are not clearly supported by the image.}
    \item The template: \textit{$<$ImageHere$>$\textcolor{blue}{the prompt}.}
\end{enumerate}
We manually verify randomly drawn parts of the instruction sets to validate the quality of generated instructions.

\begin{table*}[t]
\centering
\scalebox{0.65}{
\begin{tabular}{|l|l|l|l|l|}
\hline
Task                                  & Dataset & Split      & Type    & QA Examples     \\ \hline
\multirow{13}{*}{Scene Classification} & AID \cite{xia2017aid} & test & Optical & \multirow{11}{*}{\begin{tabular}[c]{@{}l@{}}\textbf{User}: Classify the given image in one of the classes. \\  Options: ground track field, chaparral, harbor,\\ desert, ship, railway station, meadow, bridge, ...\\ \textbf{EarthDial}: Railway station. \end{tabular}}     \\
& UCMerced-LandUse \cite{yang2010bag} & test & Optical & \\
& WHU-RS19 \cite{dai2010satellite} & test & Optical  &\\ 
 & EuroSat \cite{helber2019eurosat}  & test & Optical, S2  &\\ 
 & BigEarthNet \cite{sumbul2019bigearthnet}  & train/val/test  & Optical, S2  &  \\ 
 & NWPU-RESISC45  \cite{cheng2017remote} & train & Optical  &\\ 
 & PatternNet \cite{zhou2018patternnet} & train & Optical  &\\ 
 & RS-CD  \cite{li2020rsi} & train & Optical  &\\ 
 & RSI-CD256 \footnote{\url{https://www.kaggle.com/datasets/mahmoudreda55/satellite-image-classification/data}} & train & Optical & \\ 
 & FMoW \cite{christie2018functional} & train/val & Optical  & \\
 & FGSCR-42  \cite{di2021public}  & train & Optical & \\ 
 & TreeSatAI-Time-Series \cite{astruc2024omnisat}  & train/val/test & Optical, NIR & \\ 
& SoSAT-LCZ42 \cite{zhu2019so2sat}  & train/val/test & S2 & \\ 
 \hline

\multirow{10}{*}{Object Detection} & DOTA \cite{ding2021object} & train/test & Optical & \multirow{10}{*}{\begin{tabular}[c]{@{}l@{}}\textbf{User}:  Where is silver boeing737 airplane?\\ \textbf{EarthDial}: [bbox].\\ \textbf{User}: What object is in this location [bbox]?\\ \textbf{EarthDial}: 1 baseball field at the top right.\\ \textbf{User}: Describe this image in detail.  \\ \textbf{EarthDial}: In the image, two white motorboats  [bbox, bbox] \\  are  positioned close to each other on the left side. \end{tabular}} \\
& DOIR \cite{li2020object}  & train/test & Optical  & \\ 
 & FAIR-1M \cite{sun2022fair1m} & train/test & Optical   & \\ 
 & HIT-UAV  \cite{suo2023hit} & train/val/test & Infrared  & \\ 
 & UCAS-AOD  \cite{zhu2015orientation} & test & Optical  & \\ 
 & NWPU-VHR10 \cite{cheng2014multi}  & test & Optical  & \\ 
 & SAR-Ship \cite{wang2019sar}  & train/test & SAR  & \\ 
 & SRSDD-v1.0 \cite{lei2021srsdd} & train/test & SAR  & \\ 
 & \begin{tabular}[c]{@{}c@{}}Swimming Pool \& Car \footnote{\url{https://www.kaggle.com/datasets/kbhartiya83/swimming-pool-and-car-detection/data}}  \end{tabular} & test & Optical  & \\ 
 & Urban Tree Crown Detection \cite{zamboni2021benchmarking} & test & Optical & \\ \hline
 
\multirow{2}{*}{Visual Grounding} & DIOR-RSVG \cite{zhan2023rsvg} & train/val/test & Optical & \multirow{2}{*}{\begin{tabular}[c]{@{}l@{}}\textbf{User}: An airplane on the upper left.\\ \textbf{EarthDial}: [bbox].\end{tabular}}     \\
& RSVG \cite{sun2022visual} & train/val/test & Optical & \\  \hline
 \multirow{6}{*}{Visual Question Answering} & FloodNet \cite{rahnemoonfar2021floodnet}  & train & Optical  & \multirow{6}{*}{\begin{tabular}[c]{@{}l@{}}\textbf{User}: Is a parking present?\\ \textbf{EarthDial}: Yes.\\ \textbf{User}: What is the amount of residential buildings?\\ \textbf{EarthDial}: 14.\\ \textbf{User}: What elements in the picture suggest its theme?\\ \textbf{EarthDial}: Runway.\end{tabular}} \\
& MQVQA  \cite{zhang2023multi} & train & Optical &  \\ 
 & RSVQA-LRBEN \cite{lobry2020rsvqa} & train/test & Optical & 
 \\ 
 & RSVQA-HRBEN \cite{lobry2020rsvqa} & test & Optical &  \\ 
 & Sydney Caption \cite{qu2016deep} & train/test & Optical & 
 \\ 
 & UCM Caption \cite{qu2016deep} & train/val/test & Optical &\\ \hline
\multirow{5}{*}{Image Captioning} & RSITMD \cite{yuan2022exploring} & train/test & Optical &  \multirow{5}{*}{\begin{tabular}[c]{@{}l@{}}\textbf{User}: Could you provide the caption for input image?\\ \textbf{EarthDial}: Many white planes were parked at the airport.\\ \end{tabular}} \\
& RSCID \cite{lu2017exploring} & train/val/test & Optical & \\ 
& NWPU-Captions \cite{cheng2022nwpu} & train/val/test & Optical & \\ 
& Sydney Caption \cite{qu2016deep} & train/test & Optical &\\ 
& UCM Caption \cite{qu2016deep} & train/val/test & Optical &\\ \hline
\multirow{4}{*}{Change Detection}  & LEVIR-MCI \cite{liu2024change} & train/val/test & Optical &  \multirow{4}{*}{\begin{tabular}[c]{@{}l@{}}\textbf{User}: Are there any semantic changes detected in images?\\ \textbf{EarthDial}: Two houses are built at the top of the scene.\\ \end{tabular}} \\
& SYSU-CC \cite{noman2024cdchat} & test & Optical  & \\ 
& Dubai-CC \footnote{\url{https://disi.unitn.it/~melgani/datasets.html}} & train/val/test & Optical & \\ 
& MUDS \cite{yang2024made} & train/test & Optical & \\ \hline
\multirow{6}{*}{Methane Plume Detection}  & \multirow{6}{*}{STARCOP \cite{ruuvzivcka2023semantic}} & \multirow{6}{*}{train/test} & \multirow{6}{*}{Hyperspectral} & \multirow{6}{*}{\begin{tabular}[c]{@{}l@{}}\textbf{User}: Does this image have a methane plume?\\ \textbf{EarthDial}: Yes.\\ \textbf{User}: Give me the location of the methane plume.\\ \textbf{EarthDial}: [bbox].\\ \textbf{User}: What is the emission rate of methane plume? \\ \textbf{EarthDial}: The emission rate is 11239kg/h.\end{tabular}} \\
& & & &  \\
& & & &  \\
& & & &  \\
& & & &  \\
& & & &  \\ \hline

\multirow{6}{*}{Urban Heat Islands}  & \multirow{6}{*}{UHI-AD} & \multirow{6}{*}{train/test} & \multirow{6}{*}{Landsat8} & \multirow{6}{*}{\begin{tabular}[c]{@{}l@{}}\textbf{User}: What is the temperature trend in the input?\\ \textbf{EarthDial}: mildly hot.\\ \textbf{User}: What factors are responsible for the temperature?\\ \textbf{EarthDial}: Urbanization and few water bodies.\\ \textbf{User}:  What sustainable practices can mitigate UHI effect? \\ \textbf{EarthDial}: Introduce fountains, green corridors, and  ponds.\end{tabular}} \\
& & & &  \\
& & & &  \\
& & & &  \\
& & & &  \\
& & & & \\ \hline
\multirow{22}{*}{Disaster Assesment}  & \multirow{4}{*}{QuakeSet \cite{cambrin2024quakeset}} & \multirow{4}{*}{train/val/test} & \multirow{4}{*}{SAR} & \multirow{4}{*}{\begin{tabular}[c]{@{}l@{}}\textbf{User}: Do input images present earthquake effects?\\ \textbf{EarthDial}: Yes.\\ \textbf{User}: Could you tell the magnitude the earthquake?\\ \textbf{EarthDial}: 5.58mb.\\ \end{tabular}} \\
& & & &  \\
& & & &  \\
& & & &  \\ \cline{2-5} 
 & \multirow{18}{*}{xBD \cite{gupta2019creating}} & \multirow{18}{*}{train/test} & \multirow{18}{*}{Optical} & \multirow{18}{*}{\begin{tabular}[c]{@{}l@{}}\textbf{User}: Identify the type of disaster that occurred. \\ Options:  flood, wind, fire, tsunami, earthquake, volcano?\\ \textbf{EarthDial}: Volcano.\\ \textbf{User}: Are there any buildings affected due to disaster?\\ \textbf{EarthDial}: Yes.\\  \textbf{User}: Identify major-damaged building located at center. \\ \textbf{EarthDial}: [bbox]. \\ 
 \textbf{User}: Is the building at [bbox] affected due to disaster? \\ \textbf{EarthDial}: Yes. \\ \textbf{User}:  Describe the damage observed in the post-disaster image. \\ \textbf{EarthDial}: There has been a volcano disaster that resulted \\ in many damaged buildings. \\ \textbf{User}:  How many building are affected? \\ \textbf{EarthDial}: Many. \\ \textbf{User}:  Locate all large buildings in the post-disaster image. \\ \textbf{EarthDial}: [bbox], [bbox], [bbox].\\ \textbf{User}:  Give the level of damage for [bbox]. \\ \textbf{EarthDial}: Destroyed. \end{tabular}} \\
& & & &  \\
& & & &  \\
& & & &  \\
& & & &  \\
& & & &  \\
& & & &  \\
& & & &  \\
& & & &  \\
& & & &  \\
& & & &  \\
& & & &  \\
& & & &  \\
& & & &  \\
& & & &  \\
& & & &  \\
& & & &  \\
& & & & \\ \hline
\end{tabular}}
\caption{Overview of the downstream datasets that include various tasks, splits, types (modalities), and the generated question-answer pair  (QA-pair) examples from the respective datasets. Here, split means that we generate QA-pairs for each split separately. The [bbox] indicates the bounding box of the object as [$x_{min}, y_{min}, x_{max}, y_{max}, \theta$].}
\label{tab:list_all_downstream_datasets}
\end{table*}

\subsection*{Downstream Tasks Image-text Instruction}
\begin{figure*}[t]
\centering
 \includegraphics[width=1\linewidth]{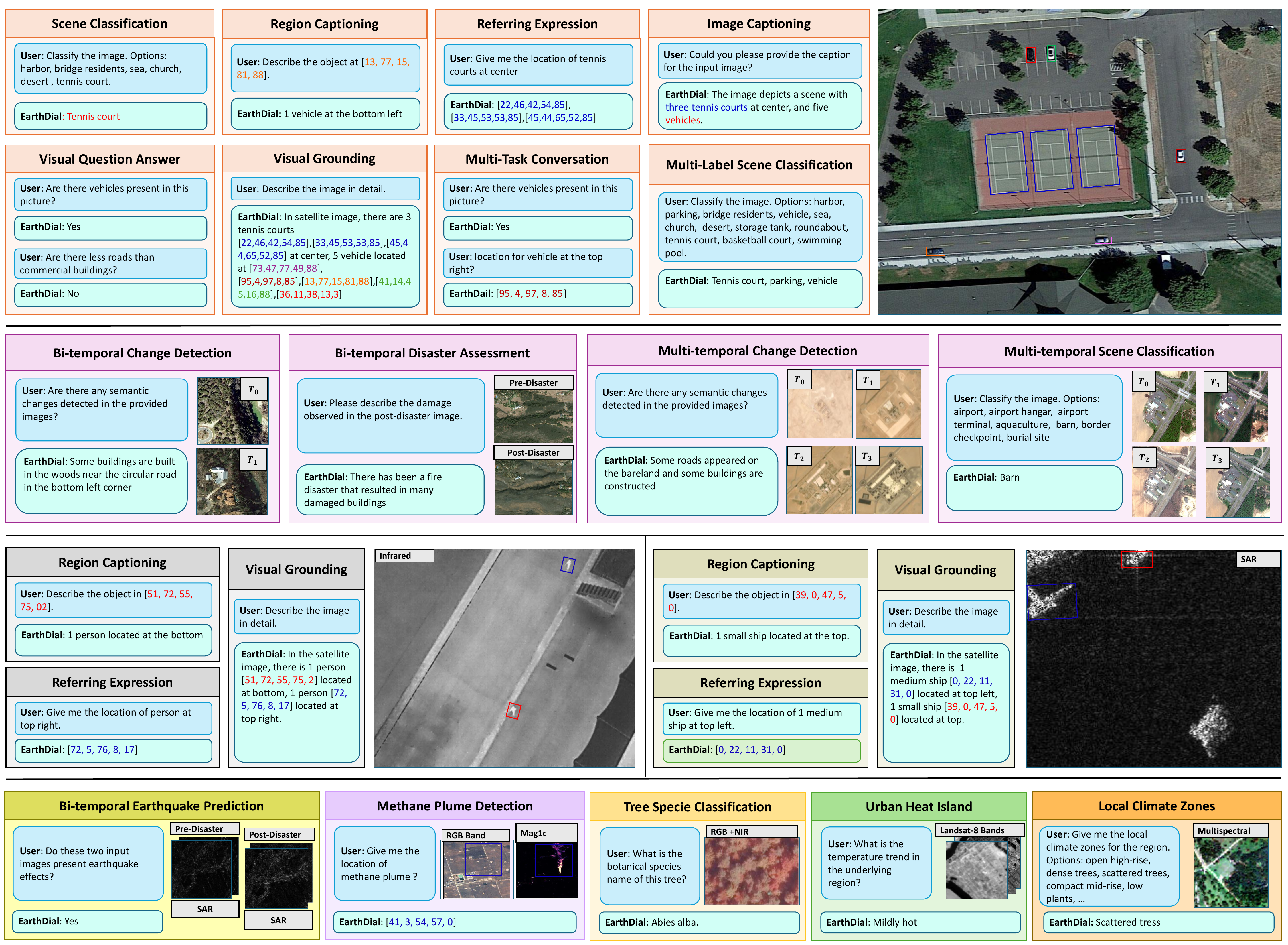} 
\caption{Illustration of our versatile \textbf{EarthDial} model that performs across multi-modalities, multi-resolution, multispectral, and multitemporal data from diverse remote sensing applications. \textbf{EarthDial} extends its capabilities to a range of tasks such as scene classification, image/region-captioning, referring expression, VQA, referring expression, object detection, temporal change/disaster detection, Methane plume detection, tree species classification, UHI, and LCZs detection across multi-modalities, multi-resolution remote sensing data.}
 \label{fig:Downstream_tasks_qa_pairs_supp}
\end{figure*}

Though pre-training enhances the generalization capabilities, we also need task-specific fine-tuning with diverse data types to improve downstream performance as shown in Tab. \ref{tab:list_all_downstream_datasets} and Fig. \ref{fig:Downstream_tasks_qa_pairs_supp}.
We curate a large number of instruction-following  datasets that include ten diverse downstream tasks: scene classification, object detection, visual question answering, image captioning,  change detection, Methane plume detection, tree species classification, local climate zones, urban heat islands, and disaster assessment. It covers seven diverse visual modalities that include Optical, SAR, S2, Infrared, NIR, Landsat8, and Hyperspectral, and two visual temporal modalities (Optical and SAR).

\begin{table*}[t]
\centering
    
    \resizebox{\textwidth}{!}{
    \begin{tabular}{l|ccccc|ccccc}
    \toprule
    \multirow{2}{*}{Model} & \multicolumn{5}{c|}{\textbf{Swimming Pool Dataset} (ZS)} & \multicolumn{5}{c}{\textbf{Urban Tree Crown Detection~\cite{zamboni2021benchmarking}} (ZS)} \\
    \cmidrule(lr){2-6} \cmidrule(lr){7-11}
     & Small & Medium & Large & Single & Multiple & Small & Medium & Large & Single & Multiple \\
    \midrule
    GeoChat~\cite{kuckreja2024geochat} & - & 3.1 & 7.3 & 1.2 & 0.6 & - & 1.8 & 8.9 & 2.9 & 3.1 \\
    InternVL2-4B~\cite{chen2024internvl} & 0.6 & 6.6 & 8.9 & 4.5 & 0.865 & -& 3.17 & 13.41 & 5.9 & 3.1 \\	
    \midrule
    \textbf{EarthDial-Lora} & \textbf{1.3} & 2.6 & 9.45 & 4.3 & 0.7 & 0.2 & 2.6 & 9.2 & 4.1 & 2.6\\
    \textbf{EarthDial (Ours)} & 1.04 & \textbf{7.4} & \textbf{24.90} & \textbf{8.4} & \textbf{1.04} & \textbf{1.1} & \textbf{7.01} & \textbf{25.67} & \textbf{11.13} & \textbf{6.7}\\
    
    \bottomrule
    \end{tabular}
    }
    \caption{Comparison of our EarthDial 
    for referred object detection tasks across various datasets. We use mAP@0.5 as the evaluation metric. Small, medium, and large denote the object size, while single and multiple denote the number of objects. Here, ZS means zero-shot evaluation.}
    \label{tab:supp_detection}
\end{table*}

\begin{figure*}[t]
\centering
 \includegraphics[width=1.0\linewidth]{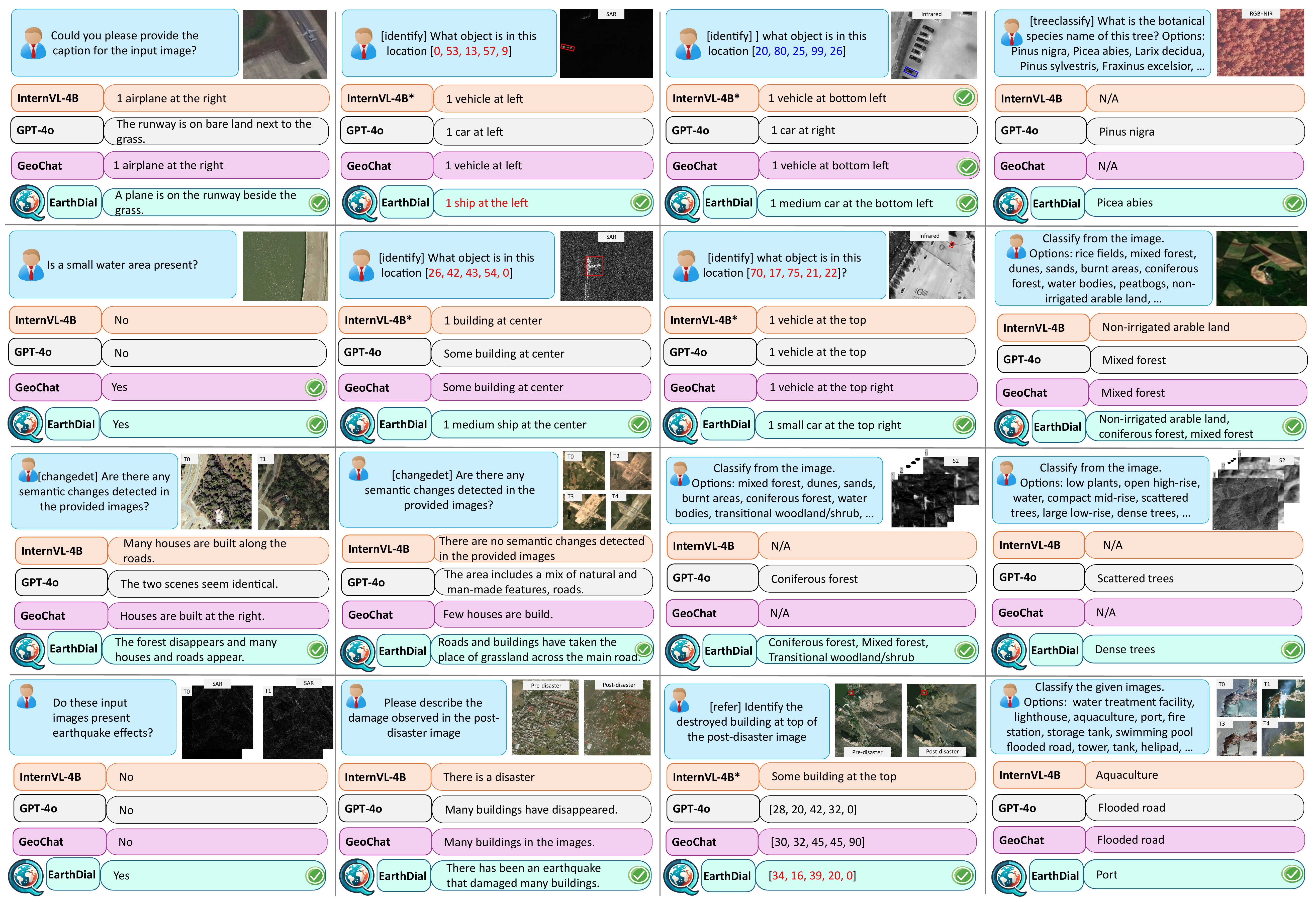} 
\caption{Illustration of the qualitative comparison of our \textbf{EarthDial} with state-of-the-art VLMs (InternVL-4B \cite{chen2024internvl}, GPT-4o \cite{openai2023gpt4}, GeoChat \cite{kuckreja2024geochat}). It demonstrates the merits of our approach by performing better under challenging scenarios across multi-modalities, multi-resolution, and temporal input data. Here, InternVL-4B* indicates that it is trained over GeoChat-Instruct. As existing InternVL2 doesn’t provide the rotated bounding boxes, for a fair comparison, we finetune the InternVL2-4B on GeoChat-Instruct and compared it with our EarthDial (only detection-related tasks).}
 \label{fig:qualitative_comparison}
\end{figure*}

\begin{figure*}[t]
\centering
 \includegraphics[width=1.0\linewidth]{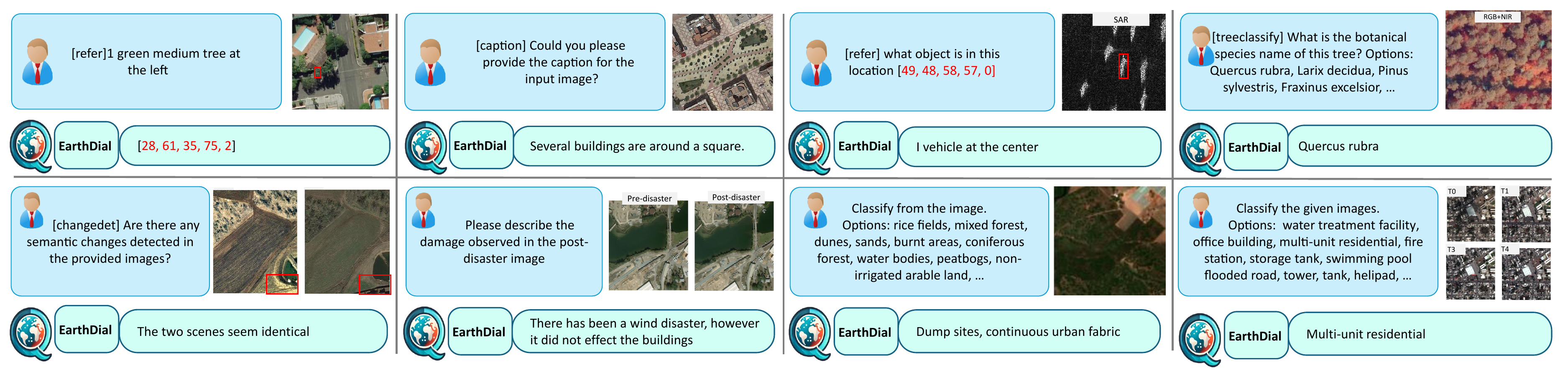} 
\caption{Illustration of the failure cases of our \textbf{EarthDial}. Our method fails under ambiguous and complex scenarios. For example, prompting the model to provide the medium tree with the input of many green trees. Similarly, for the change detection task, the model fails to detect the subtle changes that occurred at the bottom right of the scene due to variations in texture that are not easily distinguishable. }
 \label{fig:Failure_Cases}
\end{figure*}

\section{Ablation on LoRA vs Full Fine-tuning}
\label{sec:lora_exps}

It is interesting to understand how different adaptation mechanisms can influence the performance after Stage 1 model pretraining. 
Here we explore Low-rank adaptation (LoRA) in comparison to full finetuning. 
LoRA is interesting to explore since it allows finetuning the model with minimal memory requirements, adds only a few additional tunable weights and helps retain knowledge acquired during the previous training stages.
Specifically, for LoRA,  we retain the pre-trained weights from Stage 1 and instead of full finetuning, only train the low-rank adapter weights which are then added to the original pretrained weights.

For the LoRA fine-tuning, we used a LoRA rank of 128, a batch size of 2, and a learning rate of 4e-5. This setup updated approximately 201M parameters in comparison to the EarthDial model’s 4 billion total parameters while keeping the Vision Transformer (ViT), MLP, and LLM components frozen. The fine-tuning leveraged thumbnail images to capture global features and utilized an adaptive patch size ranging from 1 to 6 to capture more detailed high-level features.

The LoRA fine-tuning was performed on 2 NVIDIA A100 GPUs (80 GB each) and the model was then evaluated on zero-shot detection datasets. Compared to the fully fine-tuned model, the LoRA fine-tuned model exhibited lower performance, as summarized in Table~\ref{tab:supp_detection}. The LoRA fine-tuned model exhibited lower performance compared to the fully fine-tuned model due to its limited parameter updates, frozen components, and constrained adaptability for complex zero-shot detection tasks.

As seen from Table \ref{tab:supp_detection}, the results indicate that EarthDial (Ours) significantly outperforms EarthDial-Lora across all metrics. Specifically, EarthDial (Ours) achieves a substantial improvement in detecting multiple objects (from 2.6 to 6.7) and large objects (from 9.2 to 25.67) on the Urban Tree Crown Detection dataset. A similar trend is observed on the Swimming Pool dataset, showcasing Earthdial (full-finetunning) model's superior performance in handling the referred object detection task effectively.

    \subsection{Qualitative Analysis:}
    In Fig. \ref{fig:qualitative_comparison}, we present a qualitative analysis of EarthDial. We compare our method with existing state-of-the-art InternVL-4B ~\cite{chen2024internvl}, GPT-4o \cite{openai2023gpt4}, and GeoChat \cite{kuckreja2024geochat} VLMs.
    We notice that EarthDial shows better capability to detect the object for the SAR and infrared imagery, especially in crowded scenes.  For the multi-label scene classification, our model outputs multi-labels whereas other compared models output limits to a single label. For bi-temporal and multi-temporal change detection, we observe that our model shows better capability to identify the semantic changes in the complex scenes and indicates the newly constructed roads and buildings.
    For disaster assessment, over optical and SAR imagery, our model has better capability to identify the underlying structure and performs better for disaster understanding.
    In addition, over RGB+NIR and S2 imagery, we compare our model with GPT-4o while InternVL-4B and GeoChat do not support multi-spectral data processing. The qualitative comparison shows that our model has better capability to handle multi-spectral imagery data and performs better.
    Our qualitative comparison demonstrates the merits of EarthDial by consistently showing better performance on challenging scenarios across different modalities, multi-resolution, and multi-temporal imagery data. In Fig. \ref{fig:Failure_Cases}, we also present the failure cases where EarthDial fails under complex scenarios. For instance, identifying green medium tree at the left is difficult because there are many green trees in the input. Similarly, prompting to identify the ship provided with the bounding box may cause failure because the training set includes limited ship information compared to the vehicles. Introducing more SAR ship QA-pairs in the training set might improve the performance. On the other hand, detecting subtle change regions is difficult due to the nature of small semantic changes. For temporal scene classification, since the office building and multi-unit residential are similar in nature, therefore model might fail under such complex scenes. Nevertheless, our model encapsulates the distinctive contextual complexities of diverse RS applications and performs better compared to existing generalized and domain-specific VLMs across different modalities, multi-resolution, multi-spectral, and multi-temporal RS sensor data.

{
    \small
    \bibliographystyle{ieeenat_fullname}
    \bibliography{main}
}

\end{document}